\documentclass[final]{cvpr}
\newcommand{\name}{0}
\newcommand{\h}{0}
\newcommand{\w}{0.15}
\newcommand{\wa}{0.15}
\newlength \g
\usepackage{times}
\usepackage{epsfig}
\usepackage{graphicx}
\usepackage{amsmath}
\usepackage{amssymb}
\usepackage{multirow}
\usepackage[export]{adjustbox}

\usepackage[pagebackref=true,breaklinks=true,colorlinks,bookmarks=false]{hyperref}

\def\cvprPaperID{2435} 
\def\confYear{CVPR 2021}

\begin{document}
	
	\title{Pre-Trained Image Processing Transformer}
	
	\author{
		Hanting Chen$^{1,2}$, Yunhe Wang$^{2}$\thanks{Corresponding author}, Tianyu Guo$^{1,2}$, Chang Xu$^3$, Yiping Deng$^4$,\\
		Zhenhua Liu$^{2,5,6}$, Siwei Ma$^{5,6}$, Chunjing Xu$^2$, Chao Xu$^1$, Wen Gao$^{5,6}$\\
		\small$^1$ Key Lab of Machine Perception (MOE), Dept. of Machine Intelligence, Peking University. $^2$ Noah's Ark Lab, Huawei Technologies.\\
		\small$^3$ School of Computer Science, Faculty of Engineering, The University of Sydney. $^4$ Central Software Institution, Huawei Technologies.\\
		\small$^5$ Institute of Digital Media, School of Electronic Engineering and Computer Science, Peking University. $^6$ Peng Cheng Laboratory.\\
		\small\texttt{htchen@pku.edu.cn, yunhe.wang@huawei.com} \\ 
	}
	
	\maketitle
	\pagestyle{empty}
	\thispagestyle{empty}
	\begin{abstract}
		As the computing power of modern hardware is increasing strongly, pre-trained deep learning models (\eg, BERT, GPT-3) learned on large-scale datasets have shown their effectiveness over conventional methods. The big progress is mainly contributed to the representation ability of transformer and its variant architectures. In this paper, we study the low-level computer vision task (\eg, denoising, super-resolution and deraining) and develop a new pre-trained model, namely, image processing transformer (IPT). To maximally excavate the capability of transformer, we present to utilize the well-known ImageNet benchmark for generating a large amount of corrupted image pairs. The IPT model is trained on these images with multi-heads and multi-tails. In addition, the contrastive learning is introduced for well adapting to different image processing tasks. The pre-trained model can therefore efficiently employed on desired task after fine-tuning. With only one pre-trained model, IPT outperforms the current state-of-the-art methods on various low-level benchmarks. Code is available at \url{https://github.com/huawei-noah/Pretrained-IPT} and \url{https://gitee.com/mindspore/mindspore/tree/master/model_zoo/research/cv/IPT}
	\end{abstract}
	
	\section{Introduction}
	
	Image processing is one component of the low-level part of a more global image analysis or computer vision system. Results from the image processing can largely influence the subsequent high-level part to perform recognition and understanding of the image data. Recently, deep learning has been widely applied to solve low-level vision tasks, such as image super-resolution, inpainting, deraining and colorization. As many image processing tasks are related, it is natural to expect a model pre-trained on one dataset can be helpful for another. But few studies have generalized pre-training across image processing tasks.  
	
	Pre-training has the potential to provide an attractive solution to image processing tasks by addressing the following two challenges: First, task-specific data can be limited. This problem is exacerbated in image processing task that involves the paid-for data or data privacy, such as medical images~\cite{castellano2004texture} and satellite images~\cite{zeng2010fusion}. Various inconsistent factors (e.g. camera parameter, illumination and weather) can further perturb the distribution of the captured data for training. Second, it is unknown which type of image processing job will be requested until the test image is presented. We therefore have to prepare a series of image processing modules at hand. They have distinct aims, but some underlying operations could be shared. 
	
	\begin{figure}[t]
		\centering
		\includegraphics[width=1.0\linewidth]{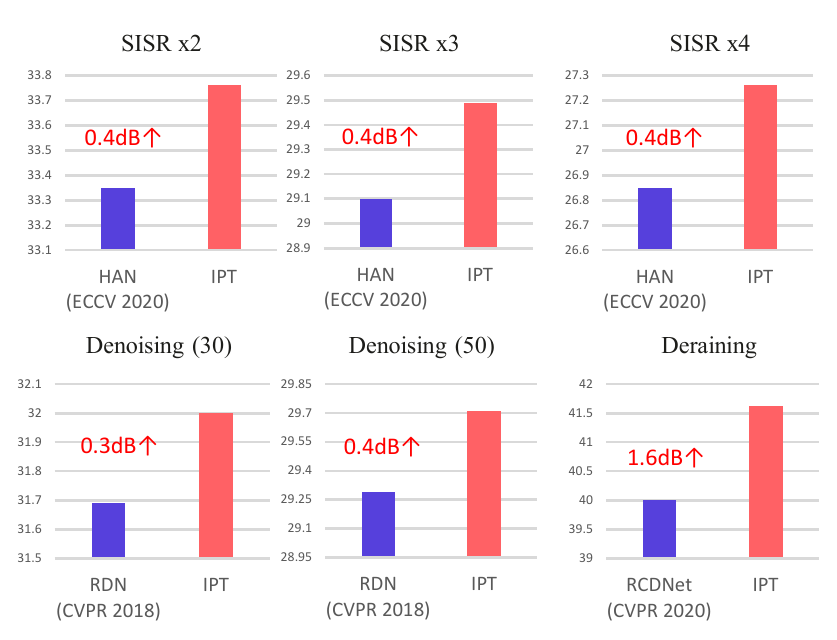}
		\caption{Comparison on the performance of the proposed IPT and the state-of-the-art image processing models on different tasks. }
		\label{fig:ipt}
		\vspace{-1.5em}
	\end{figure}
	
	It is now common to have pre-training in natural language processing and computer vision~\cite{chen2020lottery}. For example, the backbones of object detection models~\cite{zhao2019egnet,zhao2019contrast} are often pre-trained on ImageNet classification~\cite{deng2009imagenet}. A number of well-trained networks can now be easily obtained from the Internet, including AlexNet~\cite{krizhevsky2017imagenet}, VGGNet~\cite{simonyan2014very} and ResNet~\cite{he2016deep}. The seminal work Transformers~\cite{vaswani2017attention} have been widely used in many natural language processing (NLP) tasks, such as translation~\cite{wang2019learning} and question-answering~\cite{tan2019lxmert}. The secret of its success is to pre-train transformer-based models on a large text corpus and fine-tune them on the task-specific dataset. Variants of Transformers, like BERT~\cite{devlin2018bert} and GPT-3~\cite{brown2020language}, further enriched the training data and improved the pre-training skills. There have been interesting attempts on extending the success of Transformers to the computer vision field. For example, Wang~\etal~\cite{wang2017residual} and Fu~\etal~\cite{fu2019dual} applied the self-attention based models to capture global information on images. Carion~\etal~\cite{carion2020end} proposed DERT to use transformer architectures for an end-to-end object detection. Most recently, Dosovitskiy~\etal~\cite{dosovitskiy2020image} introduced Vision Transformer (ViT) to treat input images as $16\times16$ words and attained excellent results on image recognition. 
	
	The aforementioned pre-training in computer vision and natural language mostly investigate a pretest classification task, but both the input and the output in an image processing task are images. A straightforward application of these existing pre-training strategies might not be feasible. Further, how to effectively address different target image processing tasks in the pre-training stage remains a hard challenge. It is also instructive to note that the pre-training of image processing models enjoys a convenience of self-generating training instances based on the original real images. The synthetically manipulated images are taken for training, while the original image itself is the ground-truth to be reconstructed. 
	
	In this paper, we develop a pre-trained model for image processing using the transformer architecture, namely, Image Processing Transformer (IPT). As the pre-trained model needs to be compatible with different image processing tasks, including super-resolution, denoising, and deraining, the entire network is composed of multiple pairs of head and tail corresponding to different tasks and a single shared body. Since the potential of transformer needs to be excavated using large-scale dataset, we should prepair a great number of images with considerable diversity for training the IPT model. To this end, we select the ImageNet benchmark which contains various high-resolution with 1,000 categories. For each image in the ImageNet, we generate multiple corrupted counterparts using several carefully designed operations to serve different tasks. For example, training samples for the super-resolution task are generated by downsampling original images. The entired dataset we used for training IPT contains about over 10 millions of images. 
	
	Then, the transformer architecture is trained on the huge dataset as follows. The training images are input to the specific head, and the generated features are cropped into patches (\ie, ``words’’) and flattened to sequences subsequently. The transformer body is employed to process the flattened features in which position and task embedding are utilized for encoder and decoder, respectively. In addition, tails are forced to predict the original images with different output sizes according to the specific task. Moreover, a contrastive loss on the relationship between patches of different inputs is introduced for well adopting to different image processing tasks. The proposed image processing transformer is learned in an end-to-end manner. Experimental results conducted on several benchmarks show that the pre-trained IPT model can surpass most of existing methods on their own tasks by a significant enhancement after fine-tuning.

	\begin{figure*}[t]
		\centering
		\includegraphics[width=1.0\linewidth]{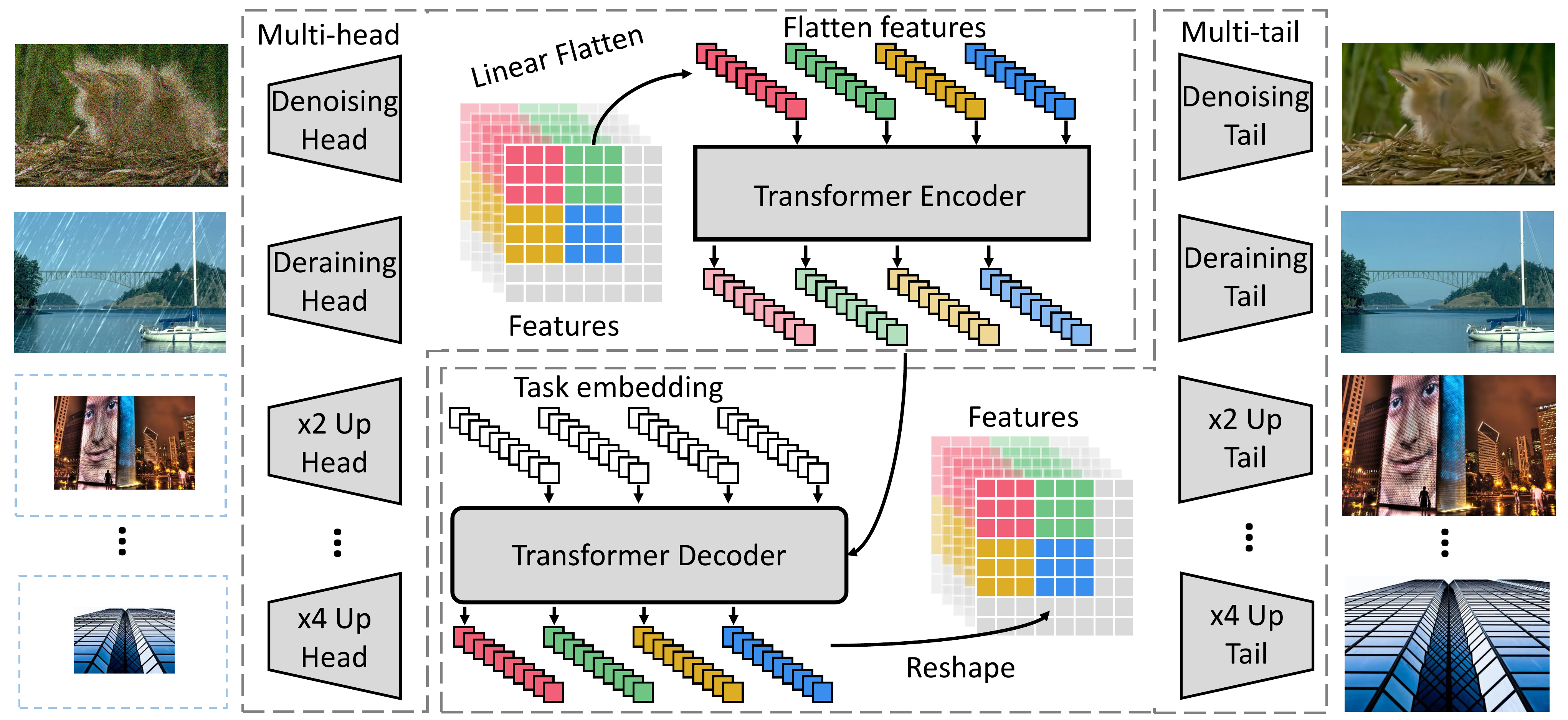}
		\caption{The diagram of the proposed image processing transformer (IPT). The IPT model consists of multi-head and multi-tail for different tasks and a shared transformer body including encoder and decoder. The input images are first converted to visual features and then divided into patches as visual words for subsequent processing. The resulting images with high visual quality are reconstructed by ensembling output patches.}
		\label{fig:intro}
		\vspace{-1.5em}
	\end{figure*}

	\section{Related Works}
	
	\subsection{Image Processing}
	
	Image processing consists of the manipulation of images, including super-resolution, denoising, dehazing, deraining, debluring, etc. There are a variety of deep-learning-based methods proposed to conduct on one or many kinds of image processing tasks. 
	For the super-resolution, Dong~\etal propose SRCNN~\cite{dong2014learning,dong2015image} which are considered as pioneering works introducing end-to-end models that reconstructs HR images from their LR counterparts. Kim~\etal~\cite{kim2016accurate} further explore the capacity of deep neural network with a more deeper convolutional network. Ahn~\etal~\cite{ahn2018fast}
	and Lim~\etal~\cite{EDSR} propose introduce residual block into SR task. Zhang~\etal~\cite{zhang2018image} and Anwar and Barnes~\cite{anwar2020densely} utilize the power of attention to enhance the performance on SR task. 
	A various excellent works are also proposed for the other tasks, such as denoising~\cite{tian2020attention,guo2019toward,jia2019focnet,lefkimmiatis2017non,fan2017balanced}, dehazing~\cite{DEHAZENET,li2017all,zhang2018densely,yang2018towards}, deraining~\cite{hu2019depth,yang2019scale,ren2019progressive,fu2019lightweight,wang2019spatial,li2021comprehensive}, and debluring~\cite{tao2018scale,lu2019unsupervised,eboli2020end,chenenhanced}. Different from above methods, we dig the capacity of both big models and huge volume of data. Then a pre-training model handling several image processing tasks is introduced.

	\subsection{Transformer}
	
	Transformer~\cite{vaswani2017attention} and its variants have proven its success being powerful unsupervised or self-supervised pre-training frameworks in various natural language processing tasks. For example, GPTs~\cite{gpt,gpt-2,brown2020language} are pre-trained in a autoregressive way that predicting next word in huge text datasets. BERT~\cite{devlin2018bert} learns from data without explicit supervision and predicts a masking word based on context. Colin~\etal~\cite{T5} proposes a universal pre-training framework for several downstream tasks. Yinhan~\etal~\cite{roberta} proposes a robust variant for original BERT. 
	
	Due to the success of Transformer-based models in the NLP field, there are many attempts to explore the benefits of Transformer in computer vision tasks. These attempts can be roughly divided into two types. The first is to introduce self-attention into the traditional convolutional neural network. Yuan~\etal~\cite{OCNET} introduce spatial attention for image segmentation. Fu~\etal~\cite{DANET} proposes DANET utilizing the context information by combining spatial and channel attention. Wang~\etal~\cite{NONLOCAL}, Chen~\etal~\cite{GloRe}, Jiang~\etal~\cite{jiang2019integral} and Zhang~\etal~\cite{LATENTCNN} also augment features by self-attention to enhance model performance on several high-level vision tasks. The other type is to replace convolutional neural network with self-attention block.  For instance, Kolesnikov~\etal~\cite{BIT} and Dosovitskiy~\cite{dosovitskiy2020image} conduct image classification with transformer block. Carion~\etal~\cite{carion2020end} and Zhu~\etal~\cite{deformableDETR} implement transformer-based models in detection. Chen~\etal~\cite{iGPT} proposes a pre-trained GPT model for generative and classification tasks. Wu~\etal~\cite{wu2020visual} and Zhao~\etal~\cite{zhao2020exploring} propose pre-training methods for teansformer-based models for image recognition task. Jiang~\etal~\cite{jiang2021transgan} propose the TransGAN to generate images using Transformer. However, few related works focus on low-level vision tasks. In this paper, we explore a universal pre-training approach for image processing tasks.
	
	\section{Image Processing Transformer}
	
	To excavate the potential use of transformer on image processing tasks for achieving better results, here we present the image processing transformer by pre-training on large-scale dataset.
	
	\subsection{IPT architecture}
	
	The overall architecture of our IPT consists of four components: heads for extracting features from the input corrupted images (\eg, images with noise and low-resolution images), an encoder-decoder transformer is established for recovering the missing information in input data, and tails are used formapping the features into restored images. Here we briefly introduce our architecture, details can be found in the supplementary material.
	
	\textbf{Heads.} To adjust different image processing task, we use a multi-head architecture to deal with each task separately,  where each head consists of three convolutional layers. Denote the input image as $x\in \mathbb{R}^{3\times H\times W}$ (3 means R, G, and B), the head generates a feature map $f_H\in \mathbb{R}^{C\times H\times W}$ with $C$ channels and same height and width (typical we use $C=64$). The calculation can be formulated as $f_H = H^i(x)$, where $H^i$ $( i = \{1,\dots,N_t\})$ denote the head for the $i$th task and $N_t$ denotes the number of tasks.
	
	\textbf{Transformer encoder.} Before input features into the transformer body, we split the given features into patches and each patch is regarded as a "word". Specifically, the features $f_H\in \mathbb{R}^{C\times H\times W}$ are reshaped into a sequence of patches, \ie, $f_{p_i}\in\mathbb{R}^{P^2\times C}, i = \{1,\dots,N\}$, where $N = \frac{HW}{P^2}$ is the number of patches (\ie, the length of sequence) and $P$ is patch size. To maintain the position information of each patch, we add learnable position encodings $E_{p_i}\in\mathbb{R}^{P^2\times C}$ for each patch of feature $f_{p_i}$ following~\cite{dosovitskiy2020image,carion2020end}, and $E_{p_i}+f_{p_i}$ will be directly input into the transformer encoder. The architecture of encoder layer is following the original structure in~\cite{vaswani2017attention}, which has a multi-head self-attention module and a feed forward network. The output of encoder $f_{E_i}\in\mathbb{R}^{P^2\times C}$ for each patch has the same size to that of the input patch $f_{p_i}$. The calculation can be formulated as
	\begin{equation}
	\begin{aligned}
	&y_0 = \left[E_{p_1}+f_{p_1},E_{p_2}+f_{p_2},\dots,E_{p_N}+f_{p_N} \right],\\
	&q_i = k_i = v_i = \mbox{LN}(y_{i-1}), \\
	&y'_i = \mbox{MSA}(q_{i},k_{i},v_{i})+y_{i-1}, \\
	&y_i = \mbox{FFN}(\mbox{LN}(y'_i))+y'_i, \quad\quad\quad\quad\quad\quad i = 1,\dots,l\\
	&\left[f_{E_1},f_{E_2},\dots,f_{E_N}\right] = y_l,
	\end{aligned}
	\end{equation}
	where $l$ denotes the number of layers in the encoder, MSA denotes the multi-head self-attention module in the conventional transformer model~\cite{vaswani2017attention}, LN denotes the layer normalization~\cite{ba2016layer} and FFN denotes the feed forward network, which contains two fully connected layers.
	
	\textbf{Transformer decoder.} The decoder also follows the same architecture and takes the output of decoder as input in the transformer body, which consists of two multi-head self-attention (MSA) layers and one feed forward network (FFN). The difference to that of the original transformer here is that we utilize a task-specific embedding as an additional input of the decoder. These task-specific embeddings $E^i_{t}\in\mathbb{R}^{P^2\times C}, i = \{1,\dots,N_t\}$ are learned to decode features for different tasks. The calculation of decoder can be formulated as: 
	\begin{equation}
	\begin{aligned}
	&z_0 = \left[f_{E_1},f_{E_2},\dots,f_{E_N}\right],\\
	&q_i = k_i = \mbox{LN}(z_{i-1}) + E_{t}, v_i = \mbox{LN}(z_{i-1}) ,\\
	&z'_i = \mbox{MSA}(q_{i},k_{i},v_{i})+z_{i-1}, \\
	&q'_i = \mbox{LN}(z'_i)+ E_{t} , k'_i= v'_i= \mbox{LN}(z_0),\\
	&z''_i = \mbox{MSA}(q'_{i},k'_{i},v'_{i})+z'_{i}, \\
	&z_i = \mbox{FFN}(\mbox{LN}(z''_i))+z''_i, \quad\quad\quad\quad\quad\quad i = 1,\dots,l\\
	&\left[f_{D_1},f_{D_2},\dots,f_{D_N}\right] = y_l,
	\end{aligned}
	\end{equation}
	where $f_{D_i}\in\mathbb{R}^{P^2\times C}$ denotes the outputs of decoder. The decoded $N$ patched features with size $P^2\times C$ are then reshaped into the features $f_D$ with size $C\times H\times W$.
	
	\textbf{Tails.} The properties of tails are same as those of heads, we use multi tails to deal with different tasks. The calculation can be formulated as $f_T = T^i(f_D)$, where $T^i$ $( i = \{1,\dots,N_t\})$ denote the head for the $i$th task and $N_t$ denotes the number of tasks. The output $f_T$ is the resulted images size of $3\times H'\times W'$ which is determined by the specific task. For example, $H'=2H, W=2W$ for a $2\times$ super-resolution task.
	
	\subsection{Pre-training on ImageNet}
	
	Besides the architecture of transformer itself, one of the key factors for successfully training an excellent transformer is that the well use of large-scale datasets. Compared with image classification, the number of available data used for image processing task is relatively small (\eg, only 2000 images on DIV2K dataset for the image super-resolution task), we propose to utilize the well-known ImageNet as the baseline dataset for pre-training our IPT model, then we generate the entire dataset for several tasks (\eg, super-resolution and denosing) as follows.
	
	As the images in the ImageNet benchmark are of high diversity, which contains over 1 million of natural images from 1,000 different categories. These images have abundant texture and color information. We first remove the semantic label and manually synthesize a variety of corrupted images from these unlabeled images with a variety of degradation models for different tasks. Note that synthesized dataset is also usually used in these image processing tasks and we use the same degeneration methods as suggested in~\cite{gu2017joint,agustsson2017ntire}. For example, super-resolution tasks often take bicubic degradation to generate low-resolution images, denoising tasks add Gaussian noise in clean images with different noise level to generate the noisy images. These synthesized images can significantly improve the performance of learned deep networks including both CNN and transformer architectures, which will be shown in the experiment part. Basically, the corrupted images are synthesized as: 
	\begin{equation}
	I_{corrupted} =  \boldsymbol{f} (I_{clean}),
	\end{equation}
	where $\boldsymbol{f}$ denotes the degradation transformation, which is depended on the specific task: for the super-resolution task, $\boldsymbol{f}_{sr}$ is exactly the bicubic interpolation; for image denoising, $\boldsymbol{f}_{noise} (I)= I+	\eta$, where $\eta$ is the additive Gaussian noise; for deraining, $\boldsymbol{f}_{rain} (I)= I+r$ in which $r$ is a hand-crafted rain streak. The loss function for learning our IPT in the supervised fashion can be formulated as: 
	\begin{equation}
	\mathcal{L}_{supervised} =  \sum_{i=1}^{N_t} L_1(\mbox{IPT}(I^i_{corrupted}),I_{clean}),
	\label{eq:suploss}
	\end{equation}
	where $L_1$ denote the conventional L1 loss for reconstructing desired images and $I^i_{corrupted}$ denote the corrupted image for task $i$, respectively. In addition, Eq.~\ref{eq:suploss} implies that the proposed framework is trained with multiple image process tasks simultaneously. Specifically, for each batch, we randomly select one task from $N_t$ supervised tasks for training and each task will be processed using the corresponding head, tail and task embedding, simultaneously. After the pre-training the IPT model, it will capture the intrinsic features and transformations for a large variety of image processing tasks thus can be further fine-tuned to apply on the desired task using the new provided dataset. Moreover, other heads and tails will be dropped for saving the computation costs and parameters in the remained head, tail and body will be updated according to the back-propagation.
	
	However, due to the variety of degradation models, we cannot synthesize images for all image processing tasks. For example, there is a wide range of possible noise levels in practice. Therefore, the generalization ability of the resulting  IPT should be further enhanced. Similar to the pre-training natural language processing models, the relationship between patches of images is also informative. The patch in image scenario can be considered as a word in natural language processing. For example, patches cropped from the same feature map are more likely to appear together, which should be embedded into similar positions. Therefore, we introduce contrastive learning~\cite{chen2020simple,he2020momentum} for learning universal features so that the pre-trained IPT model can be utilized to unseen tasks. In practice, denote the output patched features generated by IPT decoder for the given input $x_j$ as $f^j_{D_i}\in\mathbb{R}^{P^2\times C}, i = \{1,\dots,N\}$, where $x_j$ is selected from a batch of training images $X=\{x_1,x_2,\dots,x_B\}$. We aims to minimize the distance between patched features from the same images while maximize the distance between patches from different images. The loss function for contrastive learning is formulated as: 
	\begin{equation}
	\begin{aligned}
	l(f^j_{D_{i_1}},f^j_{D_{i_2}}) = -\mbox{log}  \frac{\mbox{exp}(d(f^j_{D_{i_1}},f^j_{D_{i_2}}))}{\sum_{k= 1}^B \mathbb{I}_{k\neq j} \mbox{exp}(d(f^j_{D_{i_1}},f^k_{D_{i_2}}))},\\
	\mathcal{L}_{constrastive} =  \frac{1}{BN^2} \sum_{i_1=1}^N  \sum_{i_2=1}^N \sum_{j= 1}^B l(f^j_{D_{i_1}},f^j_{D_{i_2}}) ,
	\end{aligned}
	\end{equation}
	where $d(a,b) = \frac{a^T b}{\Vert a\Vert \Vert b\Vert }$ denotes the cosine similarity. Moreover, to make fully usage of both supervised and self-supervised information, we reformulate the loss function as:
	\begin{equation}
	\mathcal{L}_{IPT} = \lambda\cdot\mathcal{L}_{contrastive}+  \mathcal{L}_{supervised}.
	\label{ipt}
	\end{equation}
	Wherein, we combine the $\lambda$-balanced contrastive loss with the supervised loss as the final objective function of IPT. Thus, the proposed transformer network trained using Eq. 6 can be effectively exploited on various existing image processing tasks. 
	
	\section{Experiments}
	
	In this section, we evaluate the performance of the proposed IPT on various image processing tasks including super-resolution and image denoising. We show that the pre-trained IPT model can achieve state-of-the-art performance on these tasks. Moreover, extensive experiments for ablation study show that the transformer-based models perform better than convolutional neural networks when using the large-scale dataset for solving the image processing problem. 
	
	\textbf{Datasets.} To obtain better pre-trained results of the IPT model, we use the well-known ImageNet dataset, which consists of over 1M color images of high diversity. The training images are cropped into $48\times48$ patches with 3 channels for training, \ie, there are over 10M patches for training the IPT model. We then generate the corrupted images with 6 types of degradation: $2\times, 3\times,4\times$ bicubic interpolation, $30,50$ noise level Gaussian noise and adding rain-streaks, respectively. For the rain-streak generation, we follow the method described in~\cite{yang2019joint}. During the test, we crop the images in the test set into $48\times48$ patches with a 10 pixels overlap. Note that the same testing strategy is also adopted for CNN based models for a fair comparison, and the resulting PSNR values of CNN models are the same as that of their baselines.
	
	\textbf{Training \& Fine-tuning.} We use 32 Nvidia NVIDIA Tesla V100 cards to train our IPT model using the conventional Adam optimizer with $\beta_1=0.9, \beta_2=0.999$ for 300 epochs on the modified ImageNet dataset. The initial learning rate is set as $5e^{-5}$ and decayed to $2e^{-5}$ in 200 epoch with 256 batch size. Since the training set consists of different tasks, we cannot input all of them in a single batch due to the expensive memory cost. Therefore, we stack a batch of images from a randomly selected task in each iteration. After pre-training on the entire synthesized dataset, we fine-tune the IPT model on the desired task (\eg, $\times 3$ single image super-resolution) for 30 epochs with a learning rate of $2e^{-5}$. Note that SRCNN~\cite{dong2014learning} also found that using ImageNet training can bring up the performance of the super-resolution task, while we propose a model fitting general low-level vision tasks.
	
	\begin{figure*}[t]

		\centering
		\scriptsize
		
		\renewcommand{\h}{0.105}
		\renewcommand{\wa}{0.12}
		\newcommand{\wb}{0.16}
		\renewcommand{\g}{-0.7mm}
		\renewcommand{\tabcolsep}{1.8pt}
		\renewcommand{\arraystretch}{1}
		\resizebox{1.00\linewidth}{!} {
			\begin{tabular}{cc}
				\begin{adjustbox}{valign=t}
					\begin{tabular}{c}
						\includegraphics[height=0.357\textwidth, width=0.42\textwidth]{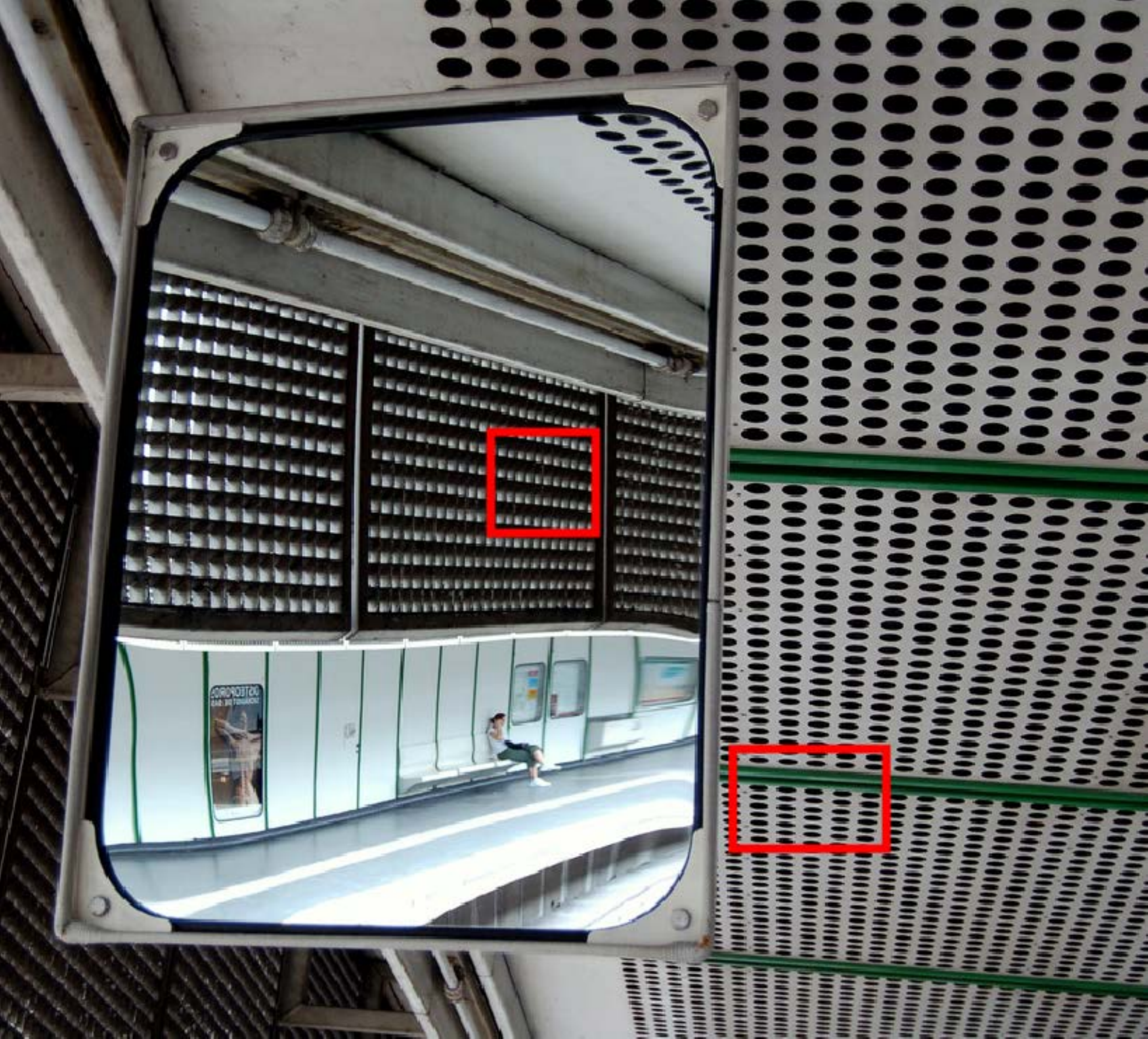}\\
						Urban100 ($\times$4): img\_004 
					\end{tabular} 
				\end{adjustbox}
				\hspace{-1.2mm}
				\begin{adjustbox}{valign=t}
					\begin{tabular}{ccc}
						\includegraphics[height=\h\textwidth, width=\wa\textwidth]{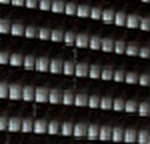} \hspace{\g}
						\includegraphics[height=\h\textwidth, width=\wb\textwidth]{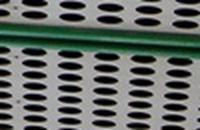} &
						\includegraphics[height=\h\textwidth, width=\wa\textwidth]{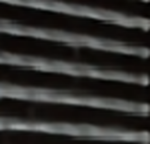} \hspace{\g} 
						\includegraphics[height=\h\textwidth, width=\wb\textwidth]{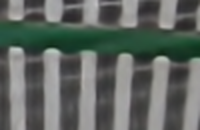} &
						\includegraphics[height=\h\textwidth, width=\wa\textwidth]{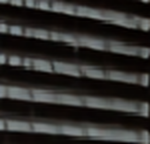} \hspace{\g}
						\includegraphics[height=\h\textwidth, width=\wb\textwidth]{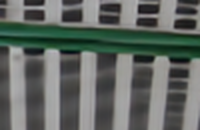} 
						
						\\
						HR &
						VDSR~\cite{kim2016accurate} &
						EDSR~\cite{lim2017enhanced} \\
						
						\includegraphics[height=\h\textwidth, width=\wa\textwidth]{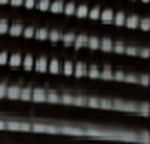} \hspace{\g}
						\includegraphics[height=\h\textwidth, width=\wb\textwidth]{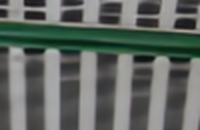} &
						\includegraphics[height=\h\textwidth, width=\wa\textwidth]{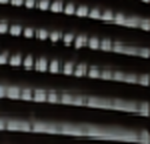} \hspace{\g} 
						\includegraphics[height=\h\textwidth, width=\wb\textwidth]{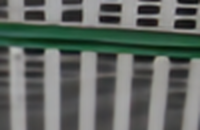} &
						\includegraphics[height=\h\textwidth, width=\wa\textwidth]{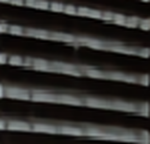} \hspace{\g}
						\includegraphics[height=\h\textwidth, width=\wb\textwidth]{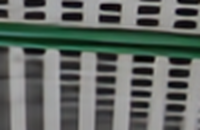}
						\\
						RDN~\cite{zhang2018residual}&
						OISR~\cite{he2019ode}  &
						SAN~\cite{dai2019second}  \\
						
						\includegraphics[height=\h\textwidth, width=\wa\textwidth]{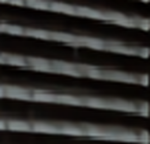} \hspace{\g}
						\includegraphics[height=\h\textwidth, width=\wb\textwidth]{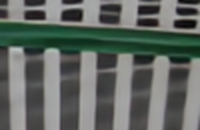} &
						\includegraphics[height=\h\textwidth, width=\wa\textwidth]{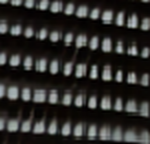} \hspace{\g} 
						\includegraphics[height=\h\textwidth, width=\wb\textwidth]{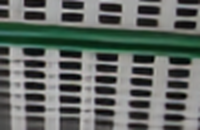} &
						\includegraphics[height=\h\textwidth, width=\wa\textwidth]{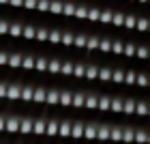} \hspace{\g} 
						\includegraphics[height=\h\textwidth, width=\wb\textwidth]{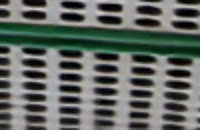} \\
						
						RNAN ~\cite{zhang2019residual} &
						IGNN~\cite{zhou2020cross}   & 
						IPT (ours)  \\
					\end{tabular} 
				\end{adjustbox}
				\\
				\\
				\renewcommand{\name}{figs/Urban_012/img_012_}
				\renewcommand{\h}{0.078}
				\renewcommand{\w}{0.176}
				\begin{tabular}{cc}
					\hspace{-2.3mm}
					\begin{adjustbox}{valign=t}
						\begin{tabular}{c}
							\includegraphics[height=0.185\textwidth, width=0.374\textwidth]{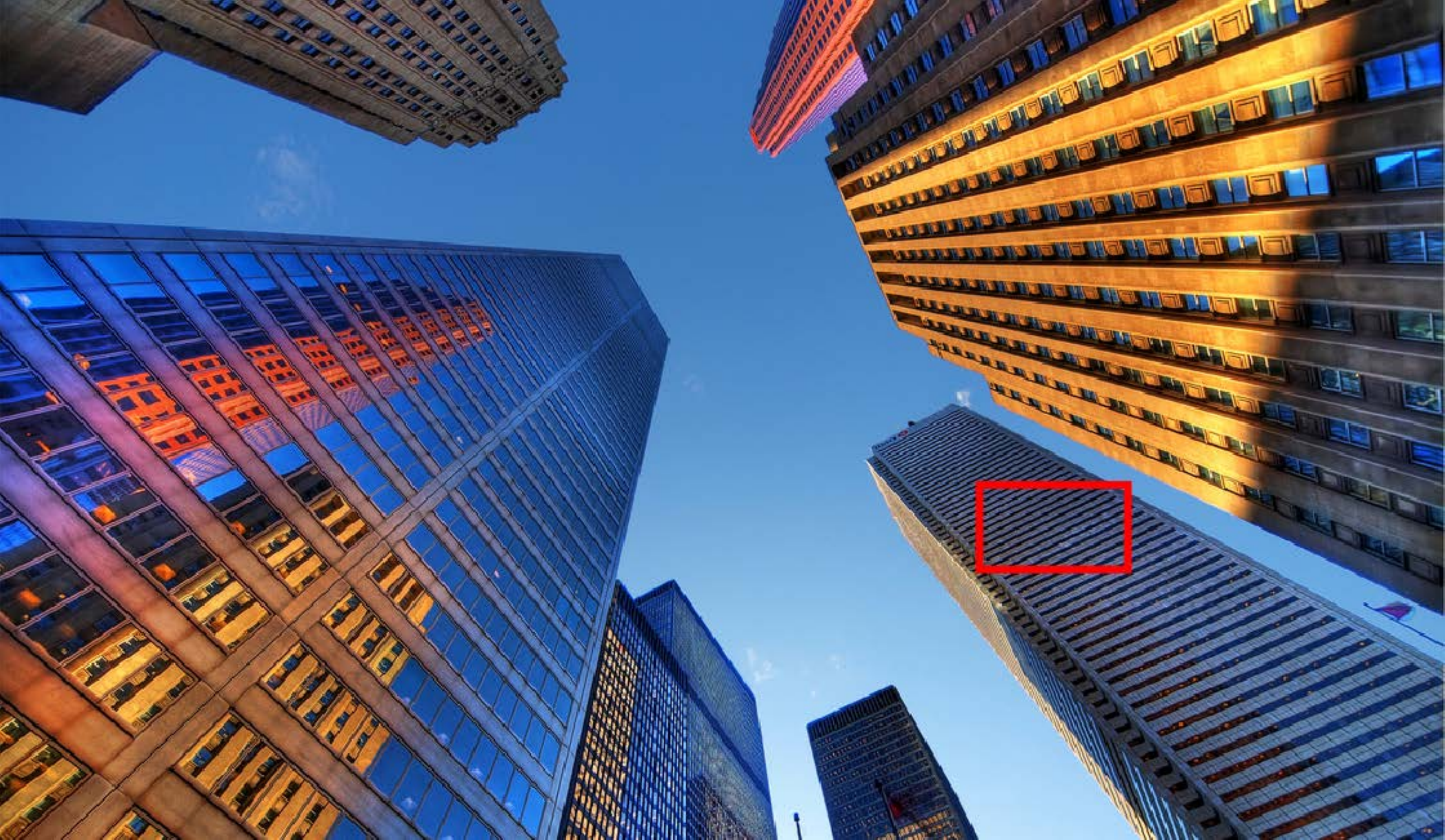}
							\\
							Urban100 ($4\times$):img\_012
						\end{tabular}
					\end{adjustbox}
					\begin{adjustbox}{valign=t}
						\begin{tabular}{cccccc}
							\includegraphics[height=\h \textwidth, width=\w \textwidth]{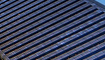} \hspace{\g} &
							\includegraphics[height=\h \textwidth, width=\w \textwidth]{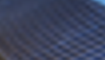} \hspace{\g} &
							\includegraphics[height=\h \textwidth, width=\w \textwidth]{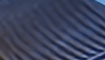} \hspace{\g} &
							\includegraphics[height=\h \textwidth, width=\w \textwidth]{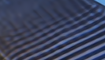} \hspace{\g} &
							\includegraphics[height=\h \textwidth, width=\w \textwidth]{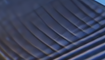} 
							\\
							HR \hspace{\g} &
							Bicubic \hspace{\g} &
							VDSR~\cite{kim2016accurate} \hspace{\g} &
							EDSR~\cite{lim2017enhanced} \hspace{\g} &
							RDN~\cite{zhang2018residual}
							\\
							\vspace{-1.5mm}
							\\
							\includegraphics[height=\h \textwidth, width=\w \textwidth]{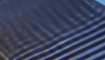} \hspace{\g} &
							\includegraphics[height=\h \textwidth, width=\w \textwidth]{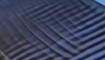} \hspace{\g} &
							\includegraphics[height=\h \textwidth, width=\w \textwidth]{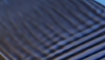} \hspace{\g} &
							\includegraphics[height=\h \textwidth, width=\w \textwidth]{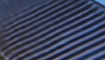}
							\hspace{\g} &		
							\includegraphics[height=\h \textwidth, width=\w \textwidth]{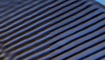}
							\\ 
							OISR~\cite{he2019ode} \hspace{\g} &
							SAN~\cite{dai2019second} \hspace{\g} &
							RNAN~\cite{zhang2019residual}  \hspace{\g} &
							IGNN~\cite{zhou2020cross} \hspace{\g} &
							IPT (ours)
							\\
						\end{tabular}
					\end{adjustbox}
				\end{tabular}
				\\
				\\
				\renewcommand{\name}{figs/Urban_044/img_044_}
				\renewcommand{\h}{0.1}
				\renewcommand{\w}{0.19}
				\begin{tabular}{cc}
					\hspace{-2.2mm}
					\begin{adjustbox}{valign=t}
						\begin{tabular}{c}
							\includegraphics[height=0.222\textwidth, width=0.304\textwidth]{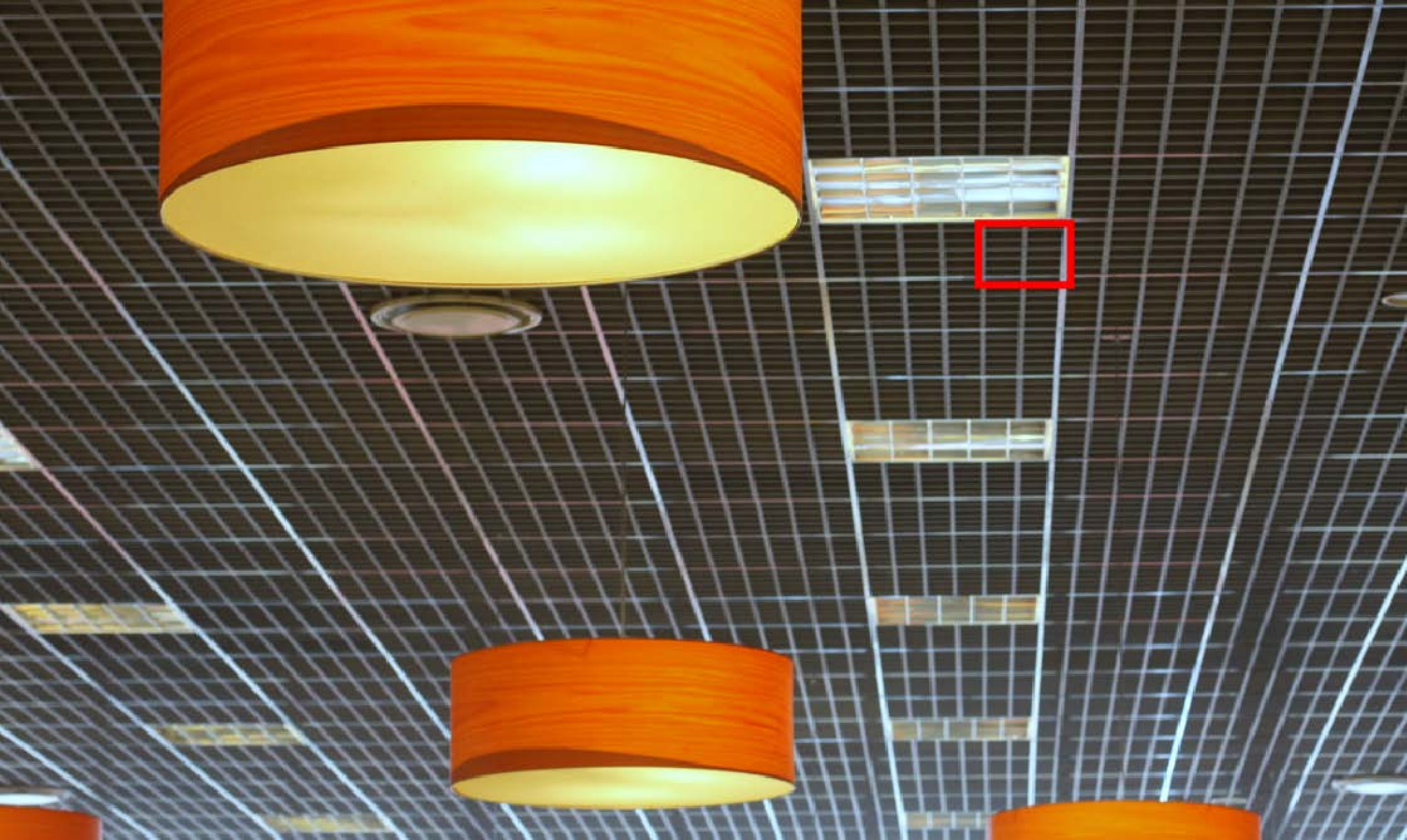}
							\\
							Urban100 ($4\times$): img\_044
						\end{tabular}
					\end{adjustbox}
					\begin{adjustbox}{valign=t}
						\begin{tabular}{cccccc}
							\includegraphics[height=\h \textwidth, width=\w \textwidth]{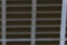} \hspace{\g} &
							\includegraphics[height=\h \textwidth, width=\w \textwidth]{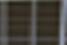} \hspace{\g} &
							\includegraphics[height=\h \textwidth, width=\w \textwidth]{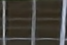} \hspace{\g} &
							\includegraphics[height=\h \textwidth, width=\w \textwidth]{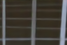} \hspace{\g} &
							\includegraphics[height=\h \textwidth, width=\w \textwidth]{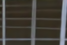} 
							\\
							HR \hspace{\g} &
							Bicubic \hspace{\g} &
							VDSR~\cite{kim2016accurate} \hspace{\g} &
							EDSR~\cite{lim2017enhanced} \hspace{\g} &
							RDN~\cite{zhang2018residual}
							\\
							\includegraphics[height=\h \textwidth, width=\w \textwidth]{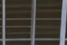} \hspace{\g} &
							\includegraphics[height=\h \textwidth, width=\w \textwidth]{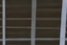} \hspace{\g} &
							\includegraphics[height=\h \textwidth, width=\w \textwidth]{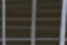} \hspace{\g} &
							\includegraphics[height=\h \textwidth, width=\w \textwidth]{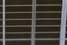} 
							\hspace{\g} &	
							\includegraphics[height=\h \textwidth, width=\w \textwidth]{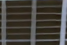}
							\\ 
							OISR~\cite{he2019ode} \hspace{\g} &
							SAN~\cite{dai2019second} \hspace{\g} &
							RNAN~\cite{zhang2019residual}  \hspace{\g} &
							IGNN~\cite{zhou2020cross} \hspace{\g} &
							IPT (ours)
							\\
						\end{tabular}
					\end{adjustbox}
				\end{tabular}
				
			\end{tabular}
		}
		\caption{Visual results with bicubic downsampling ($\times 4$) from Urban100. The proposed method recovers more details. Compared images are derived from~\cite{zhou2020cross}.}
		\label{fig:image_result}
	\end{figure*}
	
	\subsection{Super-resolution}
	
	\begin{table}[t]
		\footnotesize
		\center
		\begin{center}
			\caption{Quantitative results on image super-resolution. Best and second best results are \textbf{highlighted} and \underline{underlined}.}
			\label{tab:sr_results}
			\begin{tabular}{|l|c|c|c|c|c|}
				\hline
				Method &Scale &  Set5 &  Set14 &  B100 &  Urban100
				\\
				\hline
				\hline
				VDSR~\cite{kim2016accurate} & $\times$2 
				& 37.53
				& 33.05
				& 31.90
				& 30.77
				\\
				EDSR~\cite{lim2017enhanced} & $\times$2 
				& 38.11
				& 33.92
				& 32.32
				& 32.93
				\\
				RCAN~\cite{zhang2018image} & $\times$2 
				& {38.27}
				& {34.12}
				& {32.41}
				& {33.34}   
				\\
				RDN~\cite{zhang2018residual} & $\times$2 
				& 38.24
				& 34.01
				& 32.34
				& 32.89
				
				\\
				OISR-RK3~\cite{he2019ode} & $\times$2 
				& {38.21}
				& {33.94}
				& {32.36}
				& {33.03}   
				\\
				RNAN~\cite{zhang2019residual}& $\times$2 
				& {38.17}
				& {33.87}
				& {32.32}
				& {32.73}
				\\
				SAN~\cite{dai2019second} & $\times$2 
				& \underline{38.31}
				& {34.07}
				& \underline{32.42}
				& {33.10}
				\\
				HAN~\cite{niu2020single} & $\times$2
				&{38.27}
				& \underline{34.16}
				& {32.41}
				& \underline{33.35}
				\\
				IGNN~\cite{zhou2020cross} & $\times$2 
				& {38.24}
				& {34.07}
				& {32.41}
				& {33.23}
				\\
				\hline
				IPT (ours)  & $\times$2
				& \textbf{38.37}
				& \textbf{34.43}
				& \textbf{32.48}
				& \textbf{33.76}
				\\
				\hline                 
				\hline
				
				VDSR~\cite{kim2016accurate} & $\times$3
				& 33.67
				& 29.78 
				& 28.83 
				& 27.14
				
				\\
				EDSR~\cite{lim2017enhanced} & $\times$3 
				& 34.65
				& 30.52
				& 29.25
				& 28.80
				\\
				
				RCAN~\cite{zhang2018image}& $\times$3 
				& {34.74}
				& {30.65}
				& {29.32}
				& {29.09}
				\\
				RDN~\cite{zhang2018residual} & $\times$3 
				& 34.71
				& 30.57
				& 29.26
				& 28.80
				
				\\
				OISR-RK3~\cite{he2019ode}& $\times$3 
				& {34.72}
				& {30.57}
				& {29.29}
				& {28.95}
				\\
				RNAN~\cite{zhang2019residual}& $\times$3
				& {34.66}
				& {30.52}
				& {29.26}
				& {28.75}
				\\
				SAN~\cite{dai2019second} & $\times$3 
				&\underline{34.75}
				& {30.59}
				&\underline{29.33}
				& {28.93}
				
				\\
				HAN~\cite{niu2020single} & $\times$3 
				&\underline{34.75}
				& \underline{30.67}
				& {29.32}
				& \underline{29.10}
				
				\\
				IGNN~\cite{zhou2020cross} & $\times$3
				& {34.72}
				& {30.66}
				& {29.31}
				& {29.03}
				\\
				\hline
				IPT (ours)  & $\times$3 
				& \textbf{34.81}
				& \textbf{30.85}
				& \textbf{29.38}
				& \textbf{29.49}
				\\
				\hline
				\hline
				
				VDSR~\cite{kim2016accurate} & $\times$4
				&31.35 
				&28.02
				&27.29 
				&25.18 
				
				\\
				EDSR~\cite{lim2017enhanced} & $\times$4 
				& 32.46
				& 28.80
				& 27.71
				& 26.64
				\\
				
				RCAN~\cite{zhang2018image}& $\times$4 
				& \underline{32.63}
				& {28.87}
				& {27.77}
				&{26.82}
				
				\\
				SAN~\cite{dai2019second} & $\times$4 
				& \textbf{32.64}
				& \underline{28.92}
				& {27.78}
				& {26.79}
				\\
				
				RDN~\cite{zhang2018residual} & $\times$4 
				& 32.47
				& 28.81
				& 27.72
				& 26.61
				\\
				OISR-RK3~\cite{he2019ode} & $\times$4 
				&{32.53}
				&{28.86}
				&{27.75}
				& {26.79}
				\\
				RNAN~\cite{zhang2019residual}& $\times$4
				& {32.49}
				& {28.83}
				& {27.72}
				& {26.61}
				\\
				HAN~\cite{niu2020single}& $\times$4
				& \textbf{32.64}
				& {28.90}
				& \underline{27.80}
				& \underline{26.85}
				\\
				IGNN~\cite{zhou2020cross}  & $\times$4 
				& {32.57}
				& {28.85}
				& {27.77}
				& {26.84}
				\\
				\hline    
				IPT (ours)  & $\times$4 
				& \textbf{32.64}
				& \textbf{29.01}
				& \textbf{27.82}
				& \textbf{27.26}
				\\
				\hline  
			\end{tabular}
		\end{center}
		\vspace{-2em}
	\end{table}

	We compare our model with several state-of-the-art CNN-based SR methods. As shown in Table~\ref{tab:sr_results}, our pre-trained IPT outperforms all the other methods and achieves the best performance in $\times 2,\times3,\times4$ scale on all datasets. It is worth to highlight that our model achieves 33.76dB PSNR on the $\times2$ scale Urban100 dataset, which surpasses other methods with more than $\sim$0.4dB, while previous SOTA methods can only achieve a $<$0.2dB improvement compared with others, which indicates the superiority of the proposed model by utilizing large scale pre-training. 
	
	We further present the visualization results on our model in $4\times$ scale on Urban100 dataset. As shown in Figure~\ref{fig:image_result}, it is difficult for recover the original high resolution images since lots of information are lost due to the high scaling factor. Previous methods generated blurry images, while the super-resolution images produced by our model can well recover the details from the low-resolution images.
	
	\begin{figure*}[htpb]
		\scriptsize
		\centering
		\begin{tabular}{cc}
			\hspace{-0.4cm}
			\begin{adjustbox}{valign=t}
				\begin{tabular}{c}
					\includegraphics[width=0.2055\textwidth]{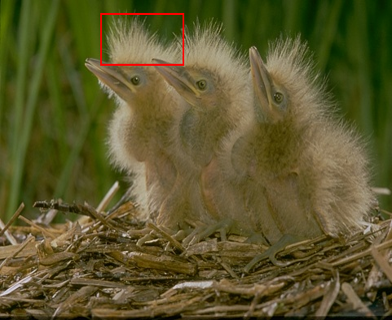}
					\\
					BSD68: 163085
				\end{tabular}
			\end{adjustbox}
			\renewcommand{\wa}{0.145}
			\renewcommand{\w}{-2.7mm}
			\hspace{-0.46cm}
			\begin{adjustbox}{valign=t}
				\begin{tabular}{cccccc}
					\includegraphics[width=\wa\textwidth]{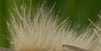} \hspace{\w} &
					\includegraphics[width=\wa\textwidth]{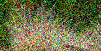} \hspace{\w} &
					\includegraphics[width=\wa\textwidth]{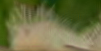} \hspace{\w} &
					\includegraphics[width=\wa\textwidth]{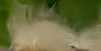} \hspace{\w} &
					\includegraphics[width=\wa\textwidth]{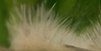} \hspace{\w} 
					\\
					GT \hspace{\w} &
					Noisy ($\sigma$=50) \hspace{\w} &
					CBM3D~\cite{dabov2007color} \hspace{\w} &
					TNRD~\cite{chen2016trainable} \hspace{\w} &
					RDN~\cite{zhang2018residual}  \hspace{\w}
					\\
					\includegraphics[width=\wa\textwidth]{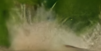} \hspace{\w} &
					\includegraphics[width=\wa\textwidth]{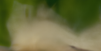} \hspace{\w} &
					\includegraphics[width=\wa\textwidth]{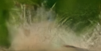} \hspace{\w} &
					\includegraphics[width=\wa\textwidth]{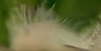} \hspace{\w} &
					\includegraphics[width=\wa\textwidth]{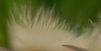} \hspace{\w}  
					\\ 
					DnCNN~\cite{zhang2017beyond} \hspace{\w} &
					MemNet~\cite{tai2017memnet} \hspace{\w} &
					IRCNN~\cite{zhang2017learning} \hspace{\w} &
					FFDNet~\cite{zhang2017ffdnet}  \hspace{\w} &
					IPT (ours) \hspace{\w}
					\\
					\\
				\end{tabular}
			\end{adjustbox}
			
		\end{tabular}
		\vspace{-2mm}
		\caption{Color image denoising results with noise level $\sigma$ = 50. Compared images are derived from~\cite{zhang2018ffdnet}.}
		\label{fig:result_DN_RGB_N50}
		\vspace{-1.5em}
	\end{figure*}

	\begin{figure*}
		\centering
		\includegraphics[width=1.0\linewidth]{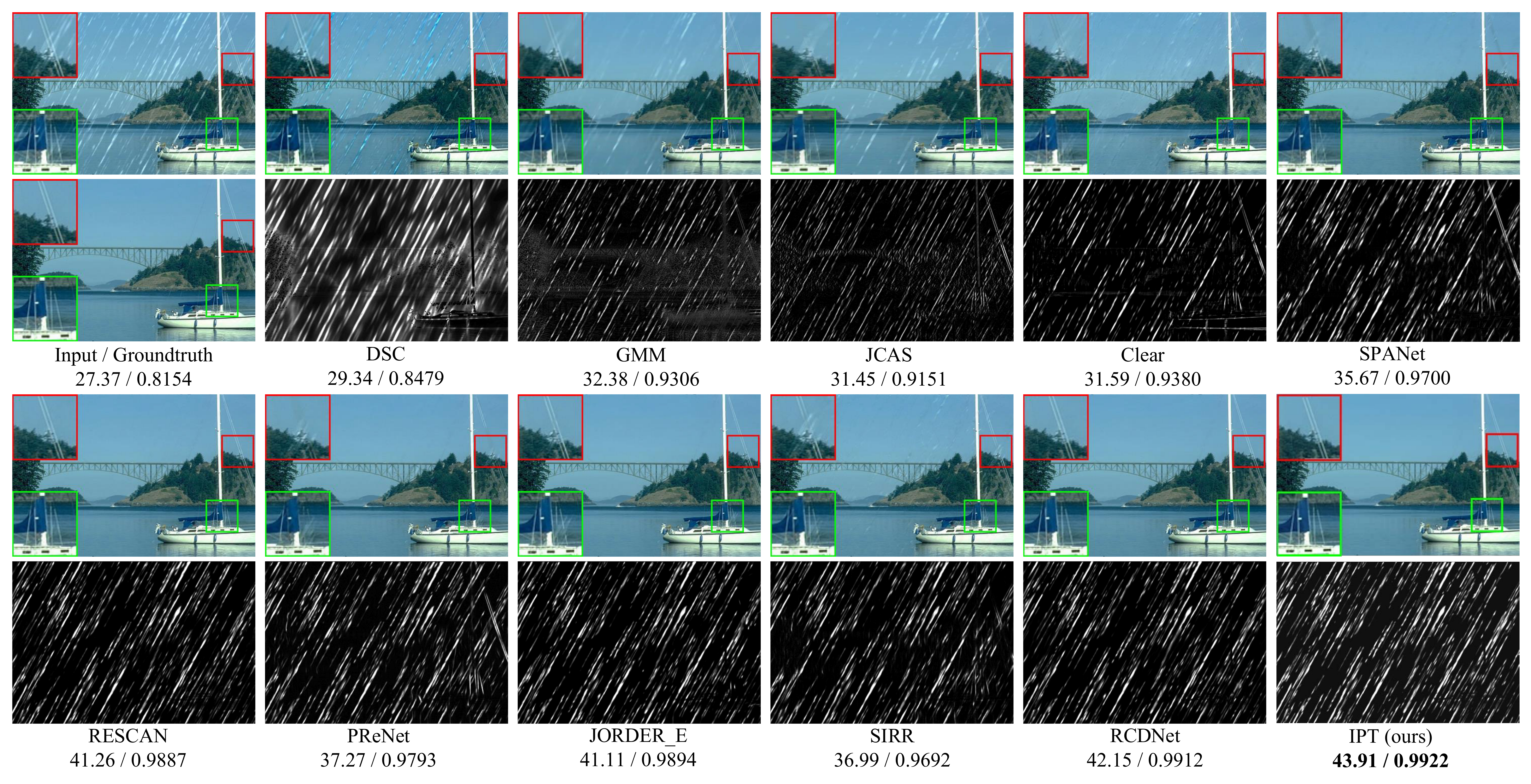}
		\caption{Image deraining results on the Rain100L dataset. Compared images are derived from~\cite{wang2020model}.}
		\label{fig:derain}
		\vspace{-1.0em}
	\end{figure*}
	\subsection{Denoising}
	
	\begin{table}[h]
		\normalsize
		\center
		\begin{center}
			\caption{Quantitative results on color image denoising. Best and second best results are \textbf{highlighted} and \underline{underlined}.}
			\label{tab:results_psnr_denoise_rgb}
			\begin{tabular}{|l|c|c|c|c|}
				\hline
				\multirow{2}{*}{Method} &   \multicolumn{2}{c|}{BSD68} &  \multicolumn{2}{c|}{Urban100}   
				\\
				\cline{2-5}
				& 30 & 50  & 30 & 50  
				\\
				\hline
				\hline
				CBM3D~\cite{dabov2007color}
				& 29.73
				& 27.38
				& 30.36
				& 27.94
				
				\\
				
				TNRD~\cite{chen2016trainable}
				& 27.64
				& 25.96
				& 27.40
				& 25.52
				
				\\

				DnCNN~\cite{zhang2017beyond}
				& 30.40
				& 28.01
				& 30.28
				& 28.16
				
				\\
				
				MemNet~\cite{tai2017memnet}
				& 28.39
				& 26.33
				& 28.93
				& 26.53
				
				\\
				
				IRCNN~\cite{zhang2017learning}
				& 30.22
				& 27.86
				& 30.28
				& 27.69
				
				\\
				
				FFDNet~\cite{zhang2017ffdnet}
				& 30.31
				& 27.96
				& 30.53
				& 28.05
				
				\\
				
				SADNet~\cite{chang2020spatial}
				& 30.64
				& \underline{28.32}
				& N/A
				&  N/A
				\\
				
				RDN~\cite{zhang2020residual}
				& \underline{30.67}
				& {28.31}
				& \underline{31.69}
				& \underline{29.29}
				
				\\
				\hline
				IPT (ours)
				& \textbf{30.75}
				& \textbf{28.39}
				& \textbf{32.00}
				& \textbf{29.71}
				
				\\
				\hline     
			\end{tabular}
		\end{center}
		\vspace{-1.5em}
	\end{table}
	
	Since our pre-trained model can be well adapt to many tasks, we then evaluate the performance of our model on image denoising task. The training and testing data is generated by adding Gaussian noise with $\sigma=30,50$ to the clean images. 
	
	To verify the effectiveness of the proposed method,  we compare our results with various state-of-the-art models. Table~\ref{tab:results_psnr_denoise_rgb} reported the color image denoising results on BSD68 and Urban100 dataset. As a result, our IPT achieves the best results among all denoising methods on different Gaussian noise level. Moreover, we surprisingly found that our model improve the state-of-the-art performance by $\sim$0.3dB on the Urban100 dataset, which demonstrate the effectiveness of pre-training and the superiority of our transformer-based model. 
	
	Figure~\ref{fig:result_DN_RGB_N50} shows the visualization of the resulted images. As shown in the figure, noisy images are hard to be recognized and it is difficult to recover the clean images. Therefore, existing methods fail to reconstruct enough details and generate abnormal pixels. As a result, our pre-trained model can well recover several details in the hair of this cat and our visual quality beats all the previous models obviously. 
	
	\subsection{Deraining}
	\begin{table*}[h]
		\center
		\begin{center}
			\caption{Quantitative results of image deraining on the Rain100L dataset. Best and second best results are \textbf{highlighted} and \underline{underlined}.}
			\label{tab:rain}
			\begin{tabular}{|l|c|c|c|c|c|c|}
				\hline
				Method & Input& DSC~\cite{fu2017removing}& GMM~\cite{li2016rain}&JCAS~\cite{gu2017joint} & Clear~\cite{fu2017clearing}& DDN~\cite{fu2017removing} \\
				\hline  
				PSNR &  26.90&27.34& 29.05&28.54&30.24&32.38\\
				\hline
				SSIM &0.8384& 0.8494&0.8717&0.8524&	0.9344&	0.9258 \\
				\hline
				\hline
				RESCAN~\cite{li2018recurrent}& PReNet~\cite{ren2019progressive} &JORDER\_E~\cite{yang2019joint}& SPANet~\cite{wang2019spatial} &SSIR~\cite{wei2019semi}& RCDNet~\cite{wang2020model}& IPT (ours)\\
				\hline  
				38.52&37.45&38.59&35.33 &32.37&\underline{40.00} &\textbf{41.62}\\
				\hline
				0.9812&0.9790&0.9834&0.9694& 0.9258 & \underline{0.9860}&\textbf{0.9880}\\
				\hline
			\end{tabular}
		\end{center}
		\vspace{-0.5em}
	\end{table*}
	
	For the image deraining task, we evaluate our model on the synthesized Rain100L dataset~\cite{yang2019joint}, which consists of 100 rainy images. Quantitative results can be viewed in Table~\ref{tab:rain}. Compared with the state-of-the-art methods, we achieve the best performance (41.62dB) with an 1.62dB improvement. 
	
	Figure~\ref{fig:derain} shows the visualization results. Previous methods are failed to reconstruct the original clean images since they lack of image prior. As a result, our IPT model can present exactly the same image as the ground-truth and surpasses all the previous algorithms in visual quality. This result substantiates the generality of the proposed model.  
	
	\subsection{Generalization Ability}
	
	\begin{table}[h]
		\normalsize
		\center
		\begin{center}
			\caption{Generation ability of our IPT model on color image denoising with different noise levels. Best and second best results are \textbf{highlighted} and \underline{underlined}.}
			\label{tab:denoise}
			\begin{tabular}{|l|c|c|c|c|}
				\hline
				\multirow{2}{*}{Method} &   \multicolumn{2}{c|}{BSD68} &  \multicolumn{2}{c|}{Urban100}   
				\\
				\cline{2-5}
				& 10 & 70  & 10 & 70  
				\\
				\hline
				\hline
				CBM3D~\cite{dabov2007color}
				& 35.91
				& 26.00
				& 36.00
				& 26.31
				
				\\
				
				TNRD~\cite{chen2016trainable}
				& 33.36
				& 23.83
				& 33.60
				& 22.63
				
				\\

				DnCNN~\cite{zhang2017beyond}
				& 36.31
				& 26.56
				& 36.21
				& 26.17
				
				\\
				
				MemNet~\cite{tai2017memnet}
				& N/A
				& 25.08
				& N/A
				& 24.96
				
				\\
				
				IRCNN~\cite{zhang2017learning}
				& 36.06
				& N/A
				& 35.81
				& N/A
				
				\\
				
				FFDNet~\cite{zhang2017ffdnet}
				& 36.14
				& 26.53
				& 35.77
				& 26.39
				
				\\
				
				RDN~\cite{zhang2020residual}
				& \underline{36.47}
				& \underline{26.85}
				& \underline{36.69}
				& \underline{27.63}
				
				\\
				\hline
				IPT (ours)
				& \textbf{36.53}
				& \textbf{26.92}
				& \textbf{36.99}
				& \textbf{27.90}
				
				\\
				\hline     
			\end{tabular}
		\end{center}
		\vspace{-0.5em}
	\end{table}
	
	Although we can generate various corrupted images, natural images are of high complexity and we cannot synthesize all possible images for pre-training the transformer model. However, a good pre-trained model should have the capacity for well adapting other tasks as those in the field of NLP. To this end, we then conduct several experiments to verify the generalization ability of our model. In practice, we test corrupted images that did not include in our synthesized ImageNet dataset, \ie, image denoising with noisy level 10 and 70, respectively. We use the heads and tails for image denoising tasks as the pre-trained model.
	
	The detailed results are shown in Table~\ref{tab:denoise}, we compare the performance of using the pre-trained IPT model and the state-of-the-art methods for image denoising. Obviously, IPT model outperforms other conventional methods, which demonstrates that the pre-trained model can capture more useful information and features from the large-scale dataset.
	
	\subsection{Ablation Study}
	
	\begin{figure}
		\centering
		\includegraphics[width=1.0\linewidth]{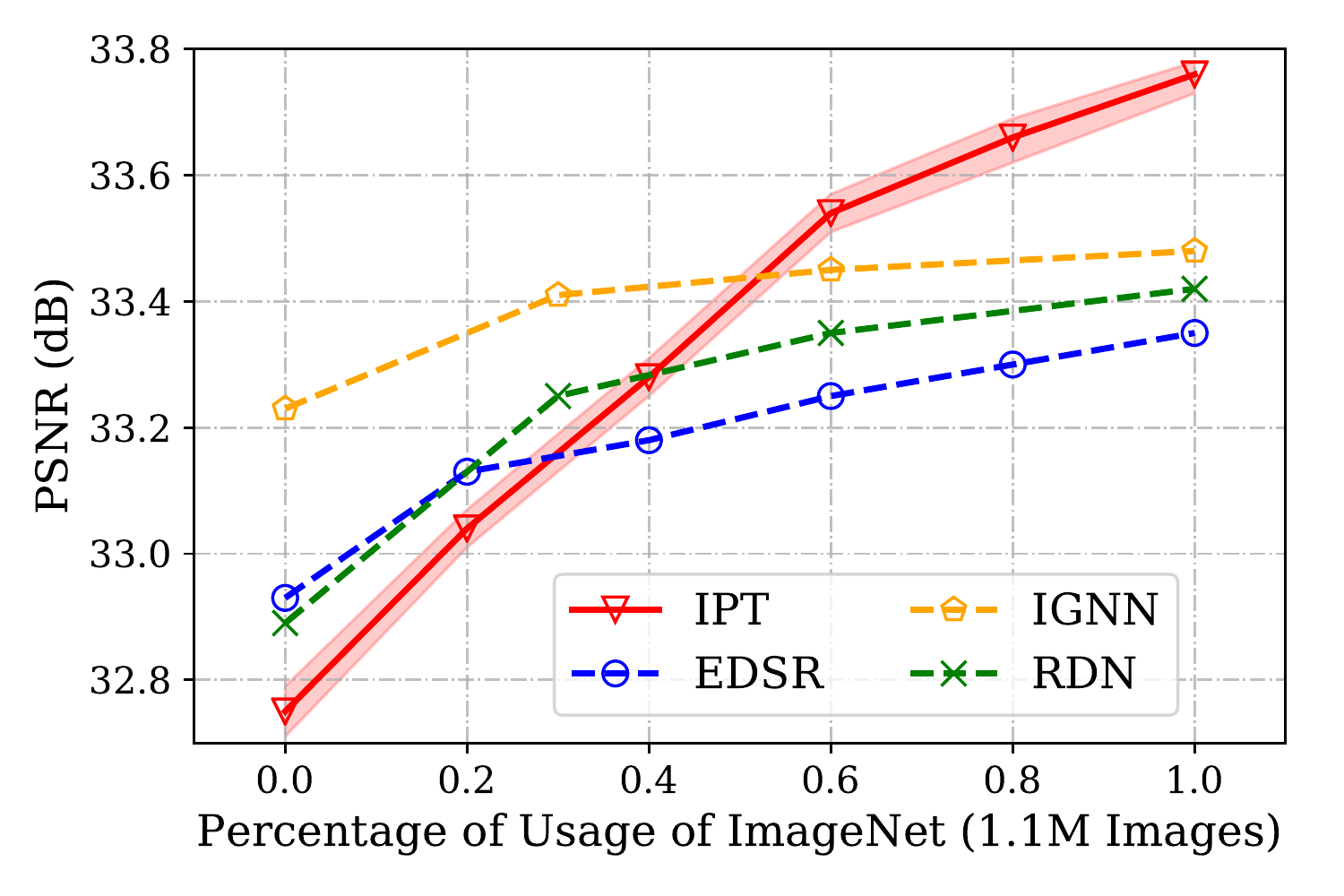}
		\caption{The performance of CNN and IPT models using different percentages of data.}
		\label{fig:abl}
		\vspace{-1.5em}
	\end{figure}
	
	\textbf{Impact of data percentage.} To evaluate the effectiveness of the transformer architecture, we conduct experiments to analyse the improvement of pre-training on CNN-based model and transformer-based model. We use 20\%, 40\%, 60\%, 80\% and 100\% percentages of the synthesized ImageNet dataset to analyse the impact on the number of used data for resulting performance. Figure~\ref{fig:abl} shows the results of different pre-trained models. When the models are not pre-trained or pre-trained with small amount ($<$ 60\%) of the entire dataset, the CNN models achieve better performance. In contrast, when using large-scale data, the transformer-based models overwhelming CNN models, which demonstrates that the effectiveness of our IPT model for pre-training.
	
	\begin{table}[h]
		\center
		\vspace{-0.5em}
		\begin{center}
			\caption{Impact of $\lambda$ for contrastive learning.}
			\label{tab:cons}
			\begin{tabular}{|l|c|c|c|c|c|}
				\hline
				$\lambda$& 0&  0.05& 0.1& 0.2& 0.5\\
				\hline
				PSNR & 38.27& 38.32& 38.37& 38.33& 38.26 \\
				\hline
			\end{tabular}
		\end{center}
		\vspace{-0.5em}
	\end{table}
	
	\textbf{Impact of contrastive learning.} As discussed above, to improve the representation ability of our pre-trained model, we embed the contrastive learning loss (Eq.~\ref{ipt}) into the training procedure. We then evaluate its effectiveness on the $\times$2 scale super-resolution task using the Set4 dataset. Table~\ref{tab:cons} shows the impact of the hyper-parameter $\lambda$ for balancing the two terms in Eq.~\ref{ipt}. When $\lambda$=0, the IPT model is trained using only a supervised learning approach, the resulting PSNR value is 38.27dB. When employing the contrastive loss for self-supervised learning, the model can achieve a 38.37dB PSNR value ($\lambda$ = 0.1), which is about 0.1dB higher than that of the model trained with $\lambda = 0$. These results further demonstrate the effectiveness of the contrastive learning for learning better pre-trained IPT model.
	
	\section{Conclusions and Discussions}
	
	This paper aims to address the image processing problems using a pre-trained transformer model (IPT). The IPT model is designed with multi-heads,multi-tails a shared transformer body for serving different image processing task such as image super-resolution and denoising. To maximally excavate the performance of the transformer architecture on various tasks, we explore a synthesized ImageNet datesets. Wherein, each original image will be degraded to a series of counterparts as paired training data. The IPT model is then trained using supervised and self-supervised approaches which shows strong ability for capturing intrinsic features for low-level image processing. Experimental results demonstrate that our IPT can outperform the state-of-the-art methods using only one pre-trained model after a quickly fine-tuning. In the future work, we will extend our IPT model to more tasks such as inpainting, dehazing, \etc.
	
	\noindent\textbf{Acknowledgment} This work is supported by National Natural Science Foundation of China under Grant No. 61876007, and Australian Research Council under Project DE180101438 and DP210101859.

	\appendix
	\section{Results on Deblurring}
	
\begin{figure*}[h]
		\centering
		\includegraphics[width=1.0\linewidth]{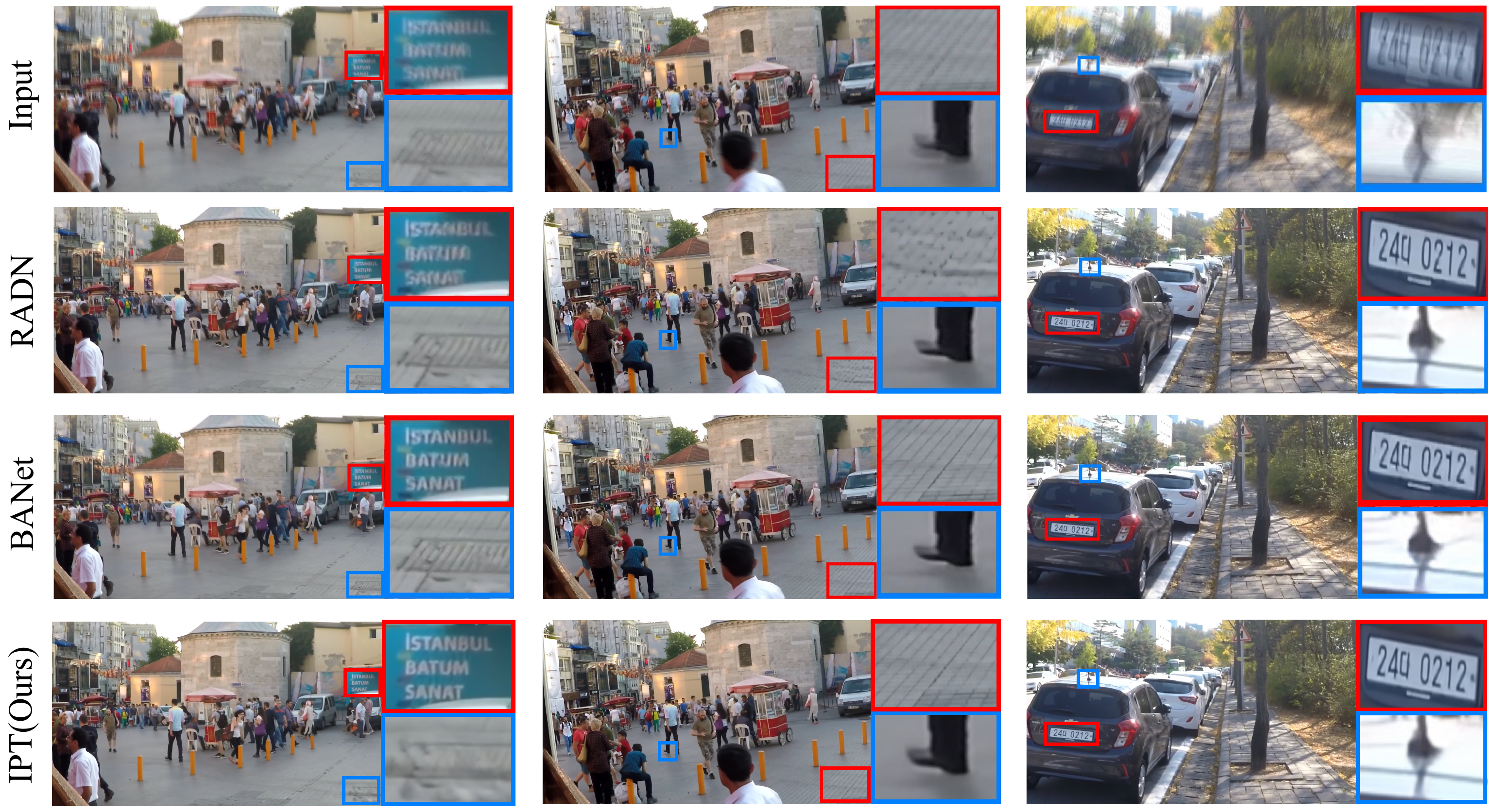}
		\caption{Image deblurring results on the GoPro dataset. Compared images are derived from~\cite{tsai2021banet}.}
		\label{fig:deblur}
	\end{figure*}

	We further evaluate the performance of our model on image deblurring task. We use the GoPro dataset~\cite{nah2017deep} to finetune and test our model. We modify the patch size as 256, patch dim as 8 and number of features as 9 to achieve a higher receptive field. Table~\ref{tab:results_psnr_deblur} reported deblurring results, where $^+$ denotes applying self-ensemble technique. As a result, our IPT achieves the best results among all deblurring methods. Figure~\ref{fig:deblur} shows the visualization of the resulted images. As shown in the figure, our pre-trained model can well achieve the best visual quality among all the previous models obviously.

	\section{Architecture of IPT}
	
	In the main paper, we propose the image processing transformer (IPT). Here we show the detailed architecture of IPT, which consists of heads, body and tails. Each head has one convolutional layer (with $3\times3$ kernel size, 3 input channels and 64 output channels) and two ResBlock. Each ResBlock consists of two convolutional layers (with $5\times5$ kernel size, 64 input channels and 64 output channels) which involved by a single shortcut. The body has 12 encoder layers and 12 decoder layers. The tail of denoising or deraining is a convolutional layer with $3\times3$ kernel size, 64 input channels and 3 output channels. For super-resolution, the tail consists of one pixelshuffle layer with upsampling scale 2 and 3 for $\times2$ and $\times3$ SR, two pixelshuffle layer with upsampling scale 2 for $\times4$ SR.
	
	The whole IPT has $114$\emph{M} parameters and $33$\emph{G} FLOPs, which have more parameters while fewer FLOPs compared with traditional CNN models (\eg, EDSR has $43$\emph{M} parameters and $99$\emph{G} FLOPs).

	\begin{figure}
		\includegraphics[width=1.0\linewidth]{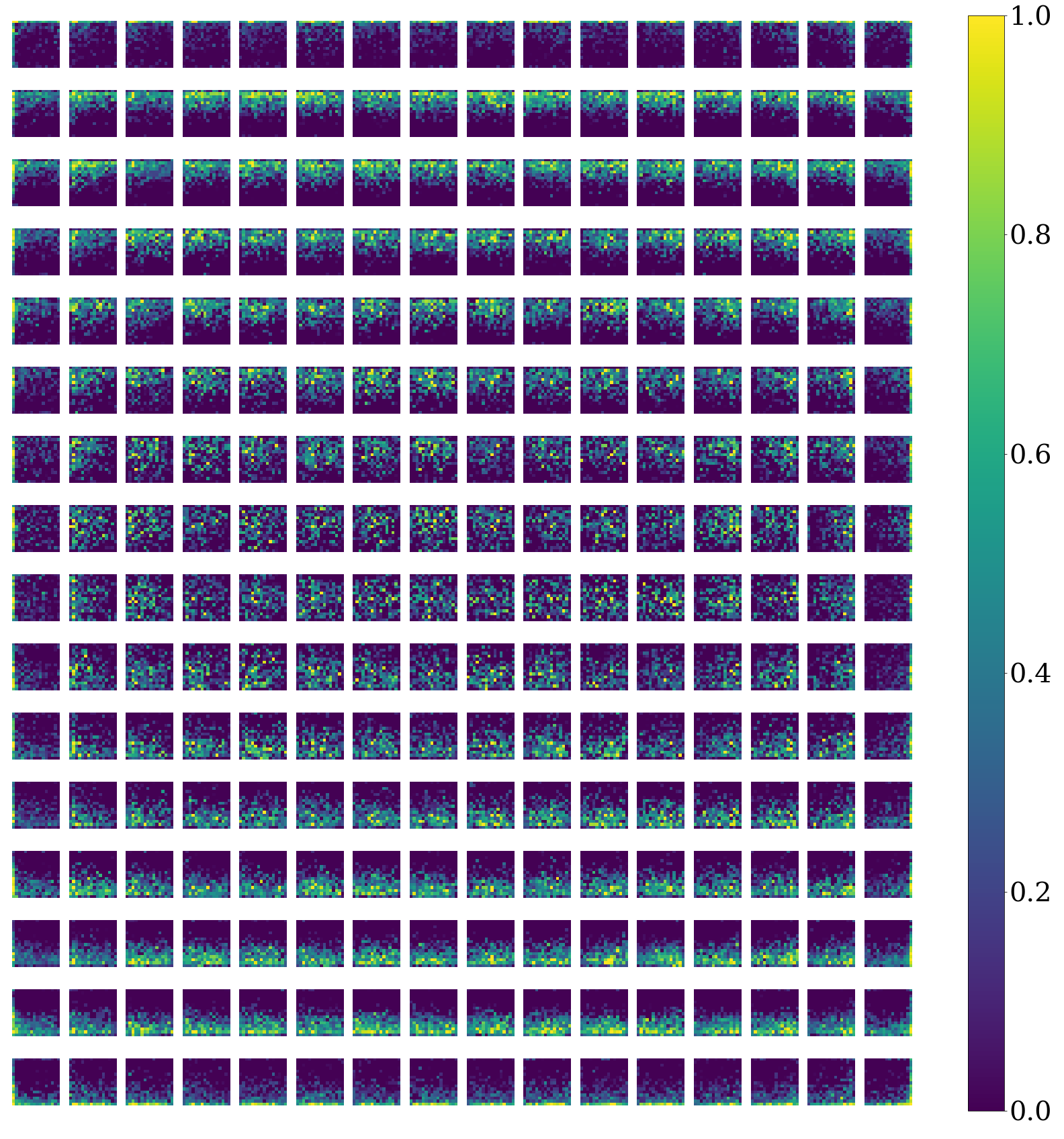}
		\caption{Visualization of cosine similarity of position embeddings.}
		\label{pos}
	\end{figure}

	\begin{table*}[h]
		\normalsize
		\center
		\begin{center}
			\caption{Quantitative results on image deblurring. Best and second best results are \textbf{highlighted} and \underline{underlined}.}
			\label{tab:results_psnr_deblur}
			\small
			\begin{tabular}{|l|c|c|c|c|c|c|c|}
				\hline
				Method & MSCNN~\cite{nah2017deep}& SRN~\cite{tao2018scale}& DSD~\cite{gao2019dynamic}& DeblurGANv2~\cite{kupyn2019deblurgan} & DMPHN~\cite{zhang2019deep}& LEBMD~\cite{jiang2020learning}&EDSD~\cite{yuan2020efficient} \\
				
				\hline  
				PSNR & 30.40 & 30.25& 30.96& 29.55& 31.36& 31.79& 29.81 \\
				\hline 
				DBGAN~\cite{zhang2020deblurring} & MTRNN~\cite{park2020multi}&RADN~\cite{purohit2020region}&SAPHN~\cite{suin2020spatially}&BANET~\cite{tsai2021banet}&	MB2D~\cite{park2020blur}&IPT (Ours) & IPT$^+$ (Ours) \\
				\hline
				31.10&31.13&31.85&32.02&32.44&32.16&\underline{32.58} & \textbf{32.91} \\
				\hline
			\end{tabular}
		\end{center}
	\end{table*}
	
	\section{Impact of Multi-task Training}
	
	We train IPT following a multi-task manner and then fine-tune it on 6 different tasks including $\times2,\times3,\times4$ super-resolution, denoising with noise level 30,50 and deraining. We find that this training strategy would not harm the performance on these tasks which have been pre-trained on large scale dataset (ImageNet). In other words, the performance of multi-task training and single-task training remains almost the same. However, when transferring to other tasks (\eg, Section 4.4 in the main paper), the pre-trained model using multi-task training is better than that of single-task training for about 0.3dB, which suggests the multi-task training would learn universal representation of image processing tasks.
	
	\section{Visualization of Embeddings}

	We visualize the learned embeddings of IPT. Figure~\ref{pos} shows the visualization results of position embeddings. We find that patches with similar columns or rows have similar embeddings, which indicate that they learn useful information for discovering the position on image processing. We also test to use fixed embeddings or do not use embeddings, whose performance are lower than that of using learnable position embeddings (vary from 0.2dB to 0.3dB for different tasks).
	
	\begin{figure*}[h]
		\centering
		\begin{tabular}{ccc}
			\includegraphics[width=0.3\linewidth]{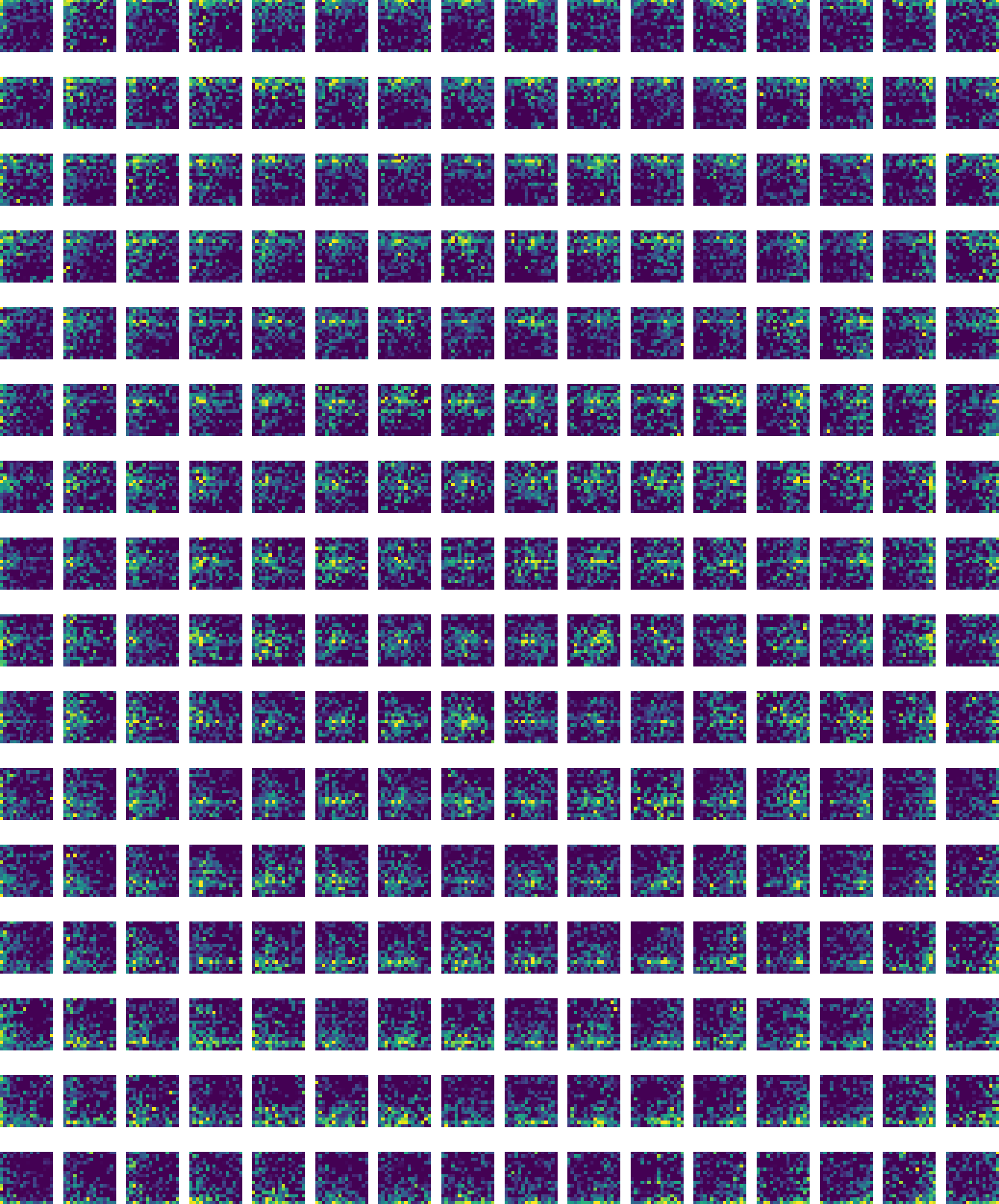} &
			\includegraphics[width=0.3\linewidth]{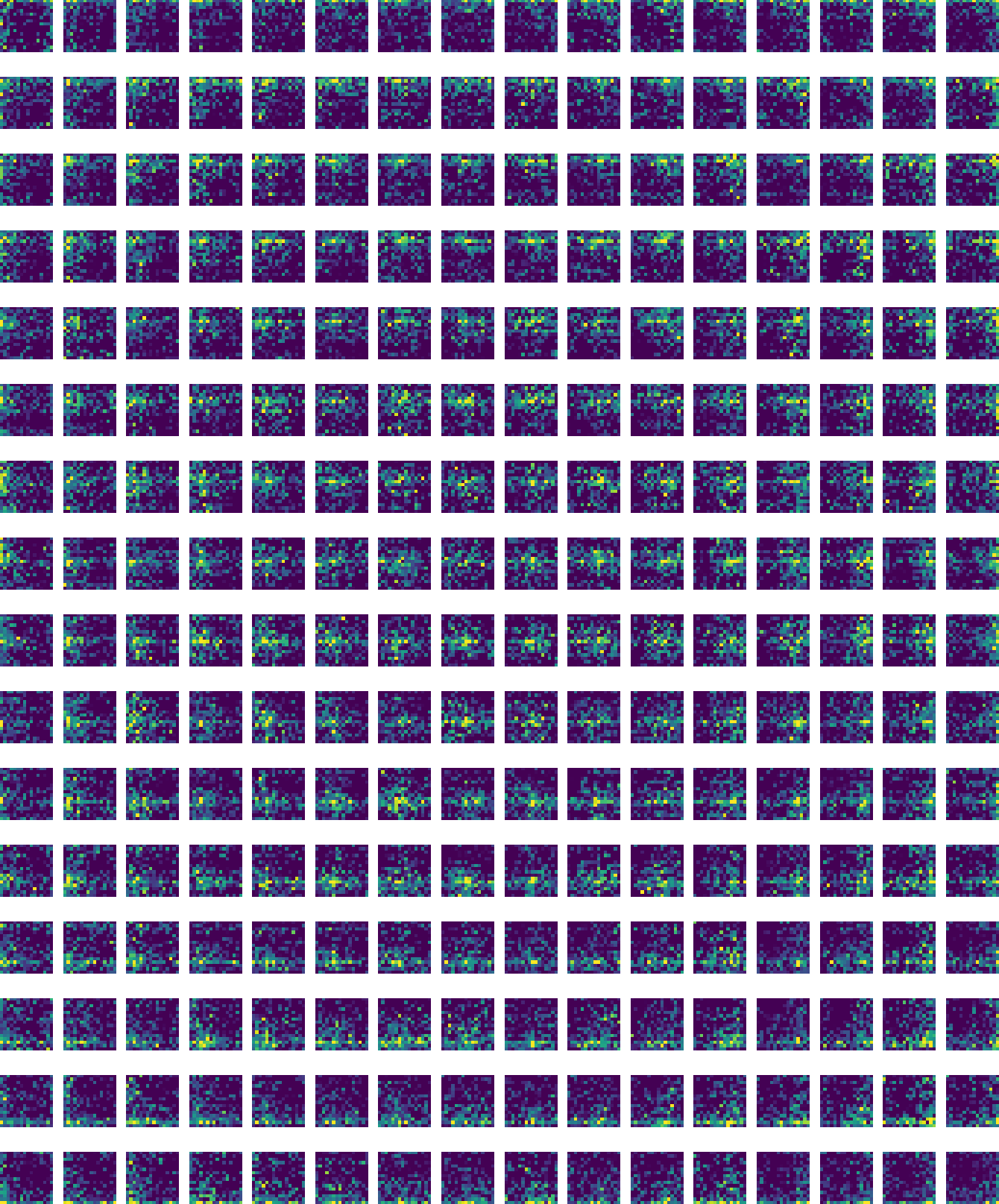}&
			\includegraphics[width=0.3\linewidth]{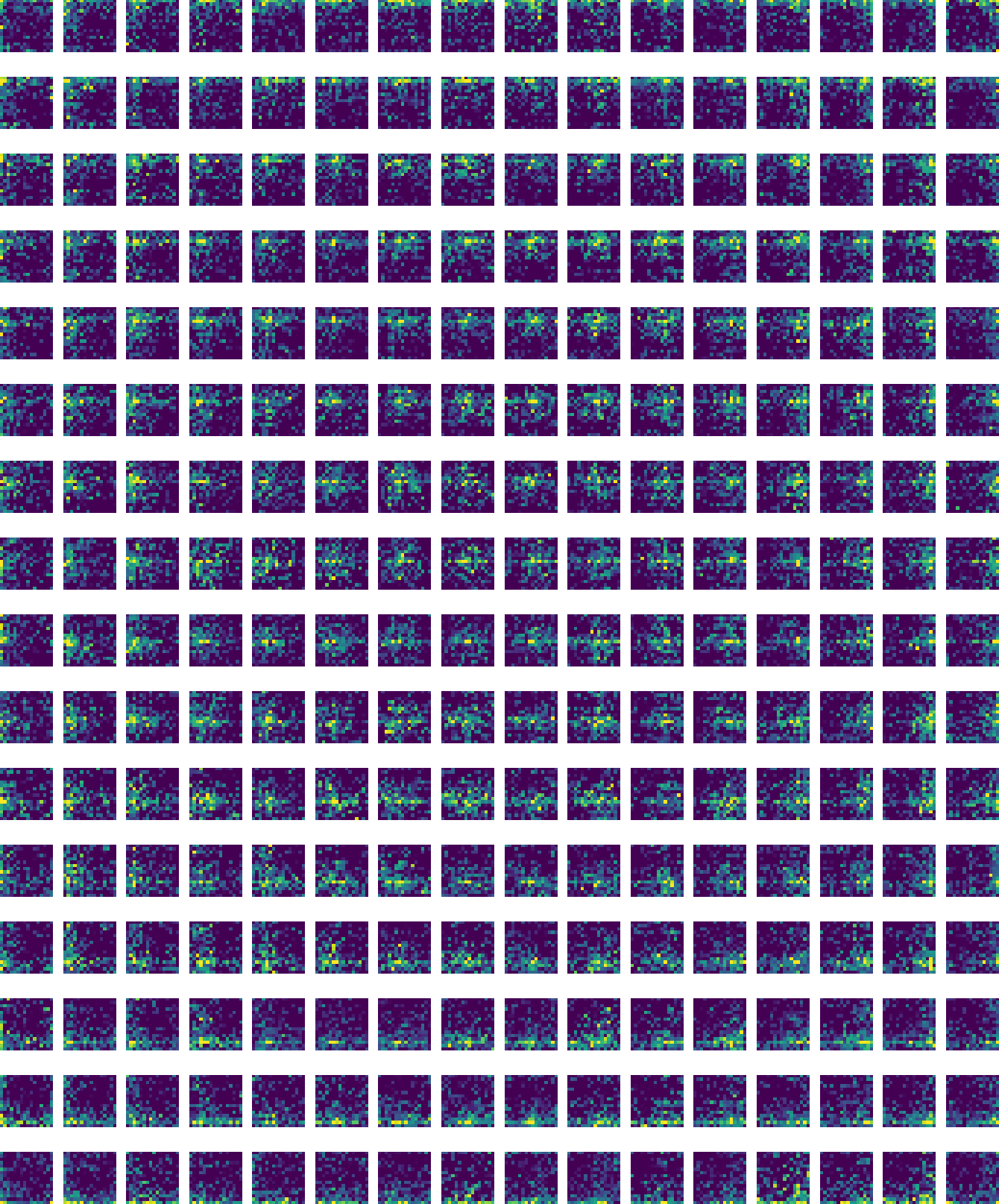}\\
			(a) $\times2$ super-resolution   &(b)  $\times3$ super-resolution & (c)  $\times4$ super-resolution  \\
			\includegraphics[width=0.3\linewidth]{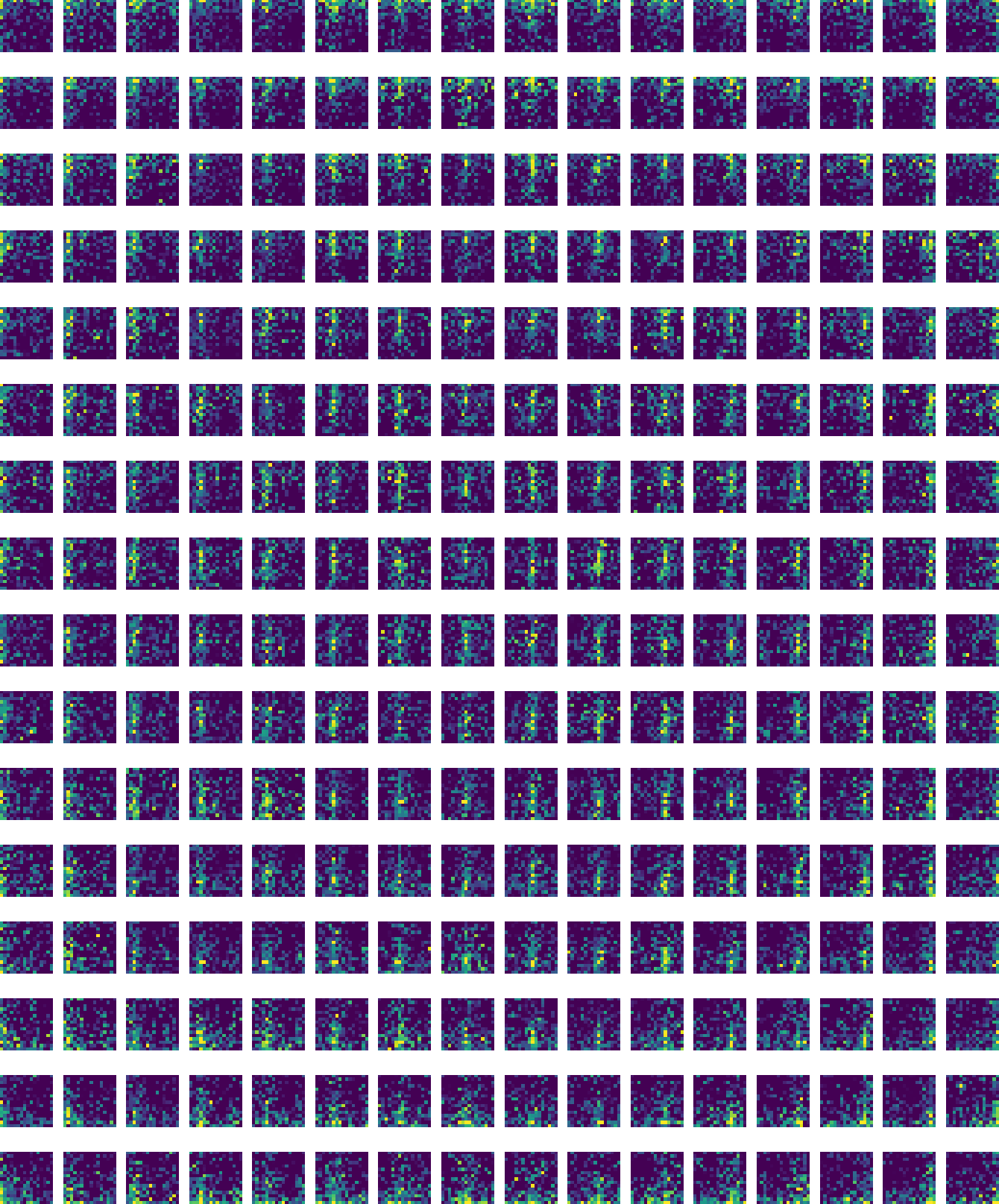} &
			\includegraphics[width=0.3\linewidth]{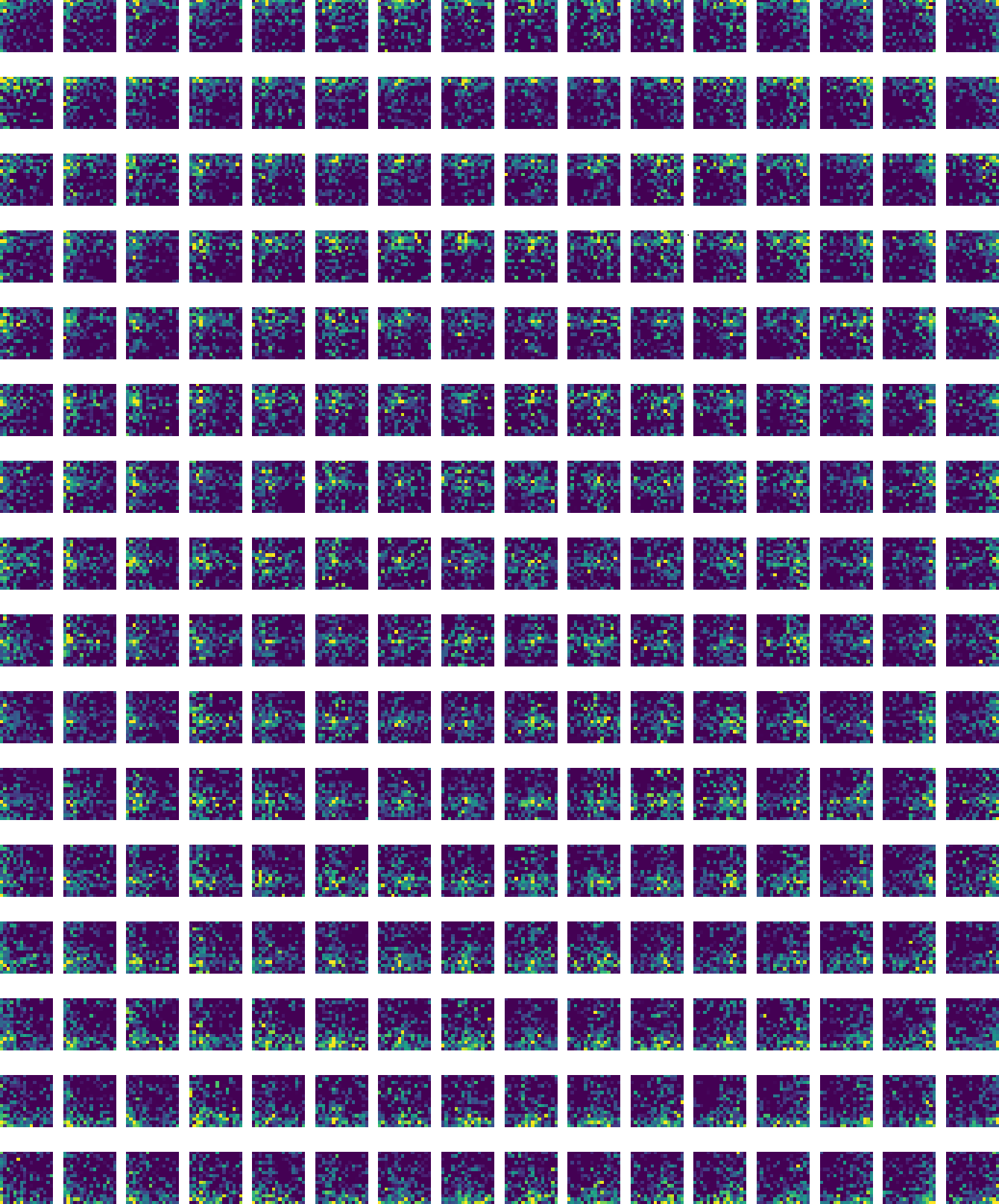}&
			\includegraphics[width=0.3\linewidth]{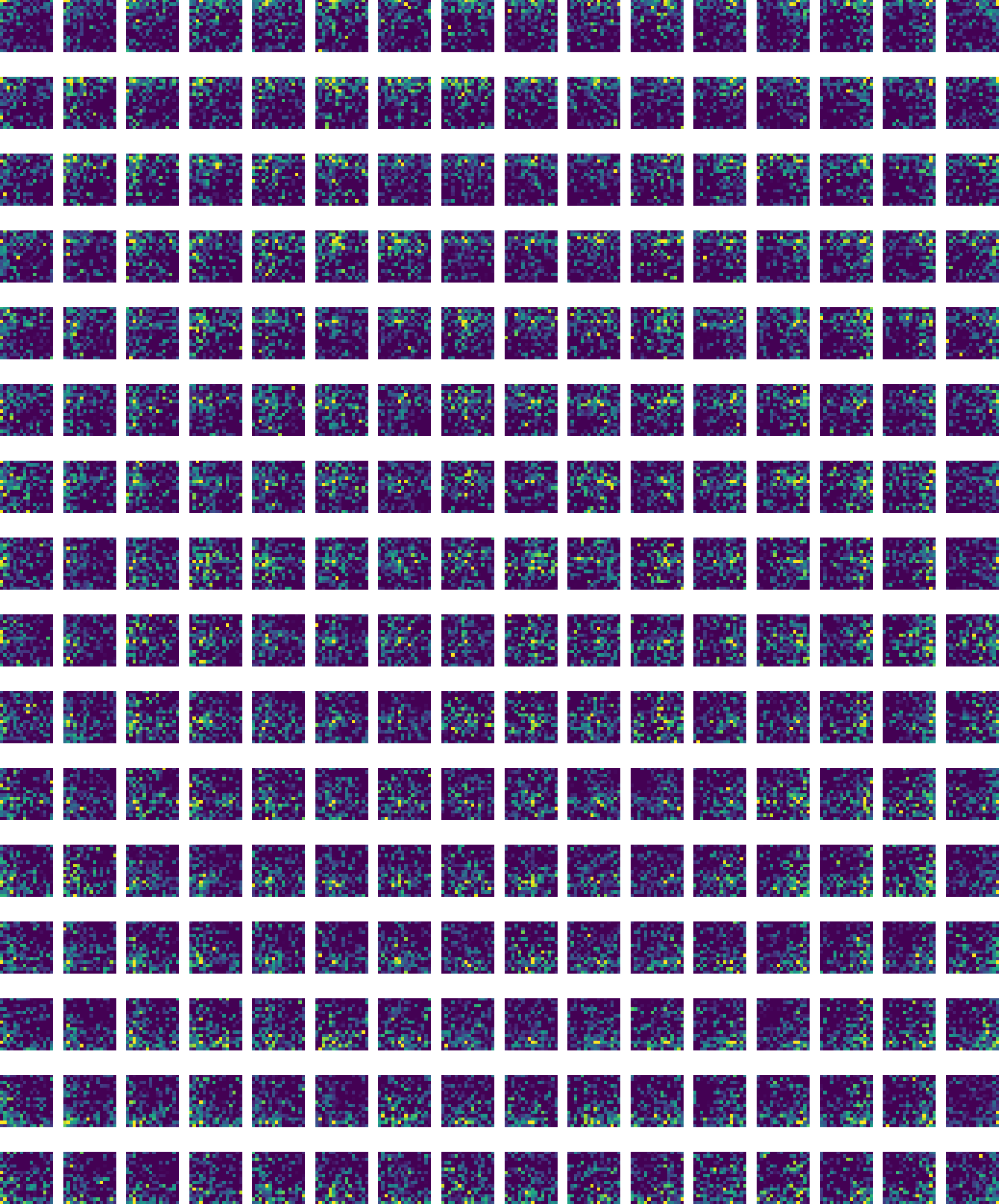}\\
			(d) deraining   &(e) denoising with 30 noise level & (f) denoising with 50 noise level  \\
			
		\end{tabular}
		\caption{Visualization of six different task embeddings.}
		\label{fig:image}
	\end{figure*}
	
	Moreover, we visualize the task embeddings in figure~\ref{fig:image}. We can find that for $\times2$ super-resolution task, the similarity between the embeddings on each position and their neighbours are higher than $\times3$ super-resolution, while that of $\times4$ super-resolution is the smallest. This results indicates that each patches in $\times2$ super-resolution can focus on other patches with farther distance than $\times3$ and $\times4$, since their downsampling scale are smaller and the relationship between different patches are closer. The similarity of task embedding for deraining in figure~\ref{fig:image} (d) shows that the patches pay more attention on the vertical direction than horizontal direction, which is reasonable as the rain is dropped vertically. The similarity of task embedding for denoising  is similar with Gaussian noise, and figure~\ref{fig:image} (f) with higher (50) noise level shows higher similarity between neighbours than figure~\ref{fig:image} (e) with 30 noise level. The visualization results suggests that our task embeddings can indeed learn some information for different tasks. We also test to not use task embeddings, which results in significant accuracy drop (vary from 0.1dB to 0.5dB for different tasks).
	
	{\small
		\bibliographystyle{ieee_fullname}
		\bibliography{egbib}
	}

\end{document}